\RequirePackage{fix-cm}
\documentclass[twocolumn]{svjour3}          
\smartqed  
\usepackage{graphicx}
\usepackage{color,xcolor}
\usepackage{amsmath}
\usepackage{amsfonts}
\usepackage{multirow}
\usepackage[misc]{ifsym}
\usepackage{subfigure}
\usepackage{colortbl}
\usepackage{soul}
\soulregister\cite7
\soulregister\ref7

\usepackage[colorlinks=true,
            linkcolor=blue,
            filecolor=blue,
            anchorcolor=blue,
            citecolor=blue]{hyperref}

\begin{document}

\title{Exploiting Inter-Sample Affinity for Knowability-Aware Universal Domain Adaptation
}

\titlerunning{Exploiting Inter-Sample Affinity for Knowability-Aware Universal Domain Adaptation} 


\author{Yifan Wang\textsuperscript{1}* \and 
        Lin Zhang\textsuperscript{1}*    \and
        Ran Song\textsuperscript{1}   \and
        Hongliang Li\textsuperscript{2}  \and 
       Paul~L.~Rosin\textsuperscript{3}   \and
        Wei Zhang\textsuperscript{1}
        }


\institute{ {\Letter}~~Ran Song (corresponding author)\vspace{-4mm} \\ 
\begin{quote}{ransong@sdu.edu.cn}\vspace{-1.5mm}\\
\\
Yifan Wang\\
{yi.fan.wang1216@gmail.com}\vspace{-1.5mm}\\
 \\
Lin Zhang\\
{zl935546110@gmail.com}\vspace{-1.5mm}\\
 \\
Hongliang Li\\
{hlli@uestc.edu.cn}\vspace{-1.5mm}\\
 \\
Paul~L.~Rosin\\
{rosinpl@cardiff.ac.uk}\vspace{-1.5mm}\\
\\
Wei Zhang \\
{davidzhang@sdu.edu.cn}\\
\\
\textsuperscript{1}\hspace{2pt}School of Control Science and Engineering, Shandong University, Jinan, China\vspace{-1.5mm}\\
 \\
\textsuperscript{2}\hspace{2pt}School of Information and Communication Engineering, University of Electronic Science and Technology of China, Chengdu, China\vspace{-1.5mm}\\
 \\
\textsuperscript{3}\hspace{2pt}School of Computer Science and Informatics, Cardiff University, Cardiff, UK\vspace{-1.5mm}\\
 \\
 \protect{*}\hspace{2pt}These authors contributed equally to this work.
 \end{quote} 
}


\date{Received: date / Accepted: date}

\maketitle

\begin{abstract}
Universal domain adaptation (UniDA) aims to transfer the knowledge of common classes from the source domain to the target domain without any prior knowledge on the label set, which requires distinguishing in the target domain the unknown samples from the known ones. Recent methods usually focused on categorizing a target sample into one of the source classes rather than distinguishing known and unknown samples, which ignores the inter-sample affinity between known and unknown samples, and may lead to suboptimal performance. Aiming at this issue, we propose a novel UniDA framework where such inter-sample affinity is exploited. Specifically, we introduce a knowability-based labeling scheme which can be divided into two steps: 1) Knowability-guided detection of known and unknown samples based on the intrinsic structure of the neighborhoods of samples, where we leverage the first singular vectors of the affinity matrix to obtain the knowability of every target sample. 2) Label refinement based on neighborhood consistency to relabel the target samples, where we refine the labels of each target sample based on its neighborhood consistency of predictions. Then, auxiliary losses based on the two steps are used to reduce the inter-sample affinity between the unknown and the known target samples. Finally, experiments on four public datasets demonstrate that our method significantly outperforms existing state-of-the-art methods.
\keywords{Domain adaptation \and Representation Learning \and Transfer Learning \and Out-of-Distribution Detection}
\end{abstract}

\section{Introduction}
\label{intro}
Unsupervised domain adaptation (UDA)~\cite{ganin2015unsupervised,saito2018maximum,long2016unsupervised,gong2013connecting,zou2018unsupervised} aims to transfer the learned knowledge from the labeled source domain to the unlabeled target domain so that the inter-sample affinities in the latter can be properly measured.

\begin{figure*}[t]
\centering
\includegraphics[width=1\linewidth]{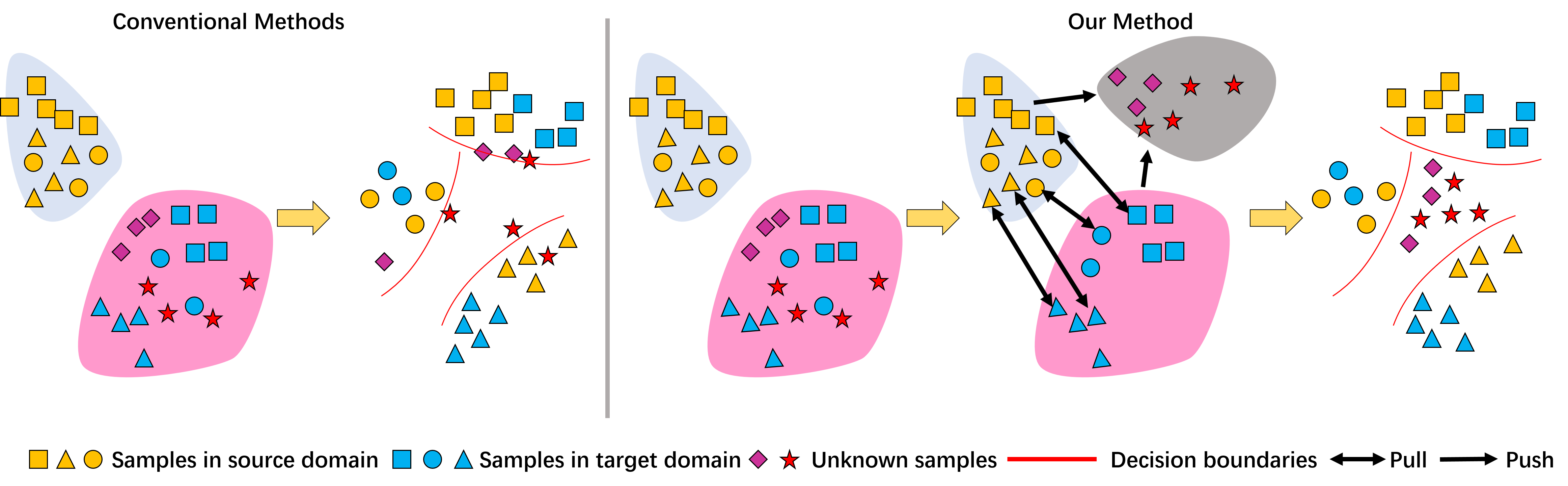}
\caption{Illustration of our method. Conventional methods usually focused on the known samples and might falsely recognise the unknown samples or ignore the inter-sample affinity between samples. Our method exploits the inter-sample affinity between known and unknown samples. The known samples in the target domain are pulled towards the corresponding samples in the source domain while the unknown samples are pushed away from any source samples.}
\label{fig:intro}
\end{figure*}

The assumption of traditional UDA, i.e., closed-set DA, is that the source domain shares an identical label set with the target domain, which significantly limits its applications in real-world scenarios. Thus, several relaxations to this assumption have been investigated. Partial-set DA~(PDA)~\cite{cao2018partial,cao2019learning,zhang2018importance,liang2020balanced} assumes that the target domain is not identical to the source domain but is a subset. On the contrary, Open-set DA~(ODA)~\cite{panareda2017open,saito2018open,liu2019separate} assumes that the target domain contains classes \emph{unknown} to the source domain such that the source domain is a subset of the target domain. Open-partial DA~(OPDA)~\cite{Saito_2021_ICCV,li2021DCC,fu2020learning} introduces private classes in both domains, where the private classes in the target domain are unknown, and assumes that the common classes shared by the two domains have been identified. 
Universal DA~(UniDA)~\cite{bucci2020effectiveness,Saito_2021_ICCV,li2021DCC} treats unsupervised DA in the most general setting, where no prior knowledge is required on the label set relationship between domains.

A popular method~\cite{you2019universal,bucci2020effectiveness,fu2020learning,Saito_2021_ICCV} for UniDA is to employ a classifier which produces a confidence for each sample to determine whether it belongs to a particular known class seen in the source domain or the unknown class.
Such methods mostly rely on the posterior probability of a classifier, which may obtain satisfactory performance on the known samples. However, as shown in the left half of Fig.~\ref{fig:intro}, once the known samples have been identified, simply ignoring unknown samples can easily lead to suboptimal classification performance for the unknown samples since such samples still contain meaningful information that can be leveraged to improve the learned representations.
In addition, the classifier-based methods may generate overconfident predictions for the known classes, leading to bias towards the known samples and the failure to identify the unknown ones.

To solve this problem, some recent approaches aim to increase the inter-sample affinity within a known class to improve the reliability of the classification.
For instance, Saito \textit{et al.} \cite{saito2020dance} proposed to assign each target sample to either a target neighbor or a prototype of a source class via entropy optimisation. Li \textit{et al.}~\cite{li2021DCC} replaced the classifier-based framework with a clustering-based one to increase the inter-sample affinity within a known class. It exploited the intrinsic structure of neighbors to directly match the clusters in the source domain and those in the target domain to discovery common and private classes. Thus, they both increased the inter-sample affinity in known classes.
However, since the inter-sample affinity between unknown samples can be greater than that between unknown and known samples due to the less discriminative features, this may lead to the misalignment between unknown samples and the prototypes in the source domain or the mismatch between the unknown clusters and the clusters in the source domain.

To mitigate such issues, we propose a novel UniDA framework which exploits the inter-sample affinity between unknown and~~~known~samples. We propose a knowability-based labeling scheme to distinguish known and unknown samples via knowability-guided detection and refine sample labels based on the neighborhood consistency of the predicted labels.~~~Specifically, the scheme can be divided into two steps: 
1) knowability-guided detection of known and unknown samples, where we decompose the affinity matrix of every target sample based on the $k$-nearest neighbors to obtain the first singular vectors as the robust representation of the local neighborhood structure and then compute the similarity between the first singular vector of each domain for every target sample to obtain the knowability;
2) label refinement based on neighborhood consistency to relabel the target samples, where each target sample is labeled via a credibility score, based on the predictions of its neighbors. Then, a target sample is labeled as known, unknown or uncertain through an automatic thresholding scheme to produce the threshold on-the-fly for the credibility score, which avoids setting the threshold manually as many existing works ~\cite{saito2020dance,li2021DCC,fu2020learning} did.

Next, we design three losses to impose a restriction on the target samples based on the above scheme. As illustrated in right half of Fig.~\ref{fig:intro}, the restriction aims to 1) reduce the inter-sample affinity between the unknown and the known samples in the target domain and 2) increase the inter-sample affinity between the known samples in the target domain and some particular samples found by the $k$-NN algorithm in the source domain where such target and source samples are supposed to belong to the same known class.

In summary, the contributions of this paper are thus fourfold:
\begin{itemize}
    \item[$\bullet$] We propose a novel method to exploit the inter-sample affinity between unknown and known samples for UniDA.

    \item[$\bullet$] We propose the knowability-guided ~detection of known and unknown samples and the label refinement based on the neighborhood consistency of each sample.

    \item[$\bullet$] We evaluate our method on four widely used UniDA benchmarks, i.e.,~ Office-31 \cite{saenko2010adapting}, OfficeHome \cite{peng2019moment}, VisDA \cite{peng2017visda} and~DomainNet \cite{venkateswara2017deep}  and~ the results demonstrate that our method considerably outperforms the state-of-the-art UniDA methods. 
\end{itemize}

\section{Related Work}
We briefly review recent unsupervised DA methods in this section. According to the assumption made about the relationship between the label sets of different domains, we group these methods into three categories, namely PDA, ODA and UniDA. We also briefly review a related problem named Out-of-Distribution Detection as it is also closely related our work.

\subsection{Partial-set Domain Adaptation}
PDA requires that the source label set is larger than and contains the target label set. Many methods for PDA have been developed \cite{cao2018san,cao2018partial,zhang2018importance,cao2019learning,liang2020balanced,liang2022dine}. For example, Cao \textit{et al.}~\cite{cao2018san} presented the selective adversarial network (SAN), which simultaneously circumvented negative transfer caused by  private source classes and promoted positive transfer between common classes in both domains to align the distributions of samples in a fine-grained manner. 
Zhang \textit{et al.}~\cite{zhang2018importance} proposed to identify common samples associated with domain similarities from the domain discriminator, and conducted a weighting operation based on such similarities for adversarial training. Cao \textit{et al.}~\cite{cao2019learning} proposed a progressive weighting scheme to estimate the transferability of source samples. Liang \textit{et al.}~\cite{liang2020balanced} introduced balanced
adversarial alignment and adaptive uncertainty suppression to avoid negative transfer and uncertainty propagation.

\subsection{Open-set Domain Adaptation}
ODA, first introduced by Busto \textit{et al.}~\cite{panareda2017open}, assumes that there are private and common classes in both source and target domains, and the labels of the common classes are known as a priori knowledge. They introduced the Assign-and-Transform-Iteratively (ATI) algorithm to address this challenging problem.

Recently, one of the most popular strategies~\cite{liu2019separate,feng2019attract} for ODA is to draw the knowledge from the domain discriminator to identify common samples across domains.
Saito~\textit{et~al.}~\cite{saito2018open} proposed an adversarial learning framework to train a classifier to obtain a boundary between source and target samples whereas the feature generator was trained to make the target samples lie far from the boundary. Bucci \textit{et al.}~\cite{bucci2020effectiveness} employed self-supervised learning technique to achieve the known/unknown separation and domain alignment.

\subsection{Universal Domain Adaptation}
UniDA, first introduced by You \textit{et al.} \cite{you2019universal} is subject to the most general setting of unsupervised DA, which involves no prior knowledge about the difference of object classes between the two domains. You \textit{et al.} also presented an universal adaptation network (UAN) to evaluate the transferability of samples based on uncertainty and domain similarity for solving the UniDA problem. However, the uncertainty and domain similarity measurements are sometimes unreliable and insufficiently discriminative. Thus, Fu \textit{et al.}~\cite{fu2020learning} proposed another transferability measure, known as Calibrated Multiple Uncertainties (CMU), estimated by a mixture of uncertainties which accurately quantified the inclination of a target sample to the common classes. 
Li \textit{et al.}~\cite{li2021DCC} introduced Domain Consensus Clustering (DCC)  to exploit the domain consensus knowledge for discovering discriminative clusters in the samples, which differentiated the unknown classes from the common ones. The latest work OVANet \cite{Saito_2021_ICCV}, proposed by Saito \textit{et al.}, trained a one-vs-all classifier for each class using labeled source samples and adapted the open-set classifier to the target domain.

\subsection{Out-of-Distribution Detection}
Out-of-Distribution (OOD) detection aims to detect OOD samples during the inference process which is enlightening to the UniDA problem of detecting unknown samples. Hendrycks \textit{et al.}~\cite{hendrycks2016baseline} first proposed a baseline method for detecting OOD samples using the confidence of classification. Recently, some methods~\cite{liang2017enhancing,lee2018simple,sastry2020detecting,hsu2020generalized} built advanced detectors in a post-hoc manner. For example,  Lee \textit{et al.}~\cite{lee2018simple} utilised the Mahalanobis distance between the features of test and the train samples
to obtain the confidence score with respect to the closest class
conditional distribution. However, these methods require many labeled samples for training. To better exploit the unlabeled data for OOD detection, Hendrycks \textit{et al.} \cite{hendrycks2018deep} enforced the model to produce the low confidence output on the pure unlabeled OOD data. Some other works~\cite{golan2018deep,hendrycks2019using,winkens2020contrastive,tack2020csi,sehwag2021ssd} employed self-supervised learning on the pure unlabeled data to improve the performance. For instance, Sehwag \textit{et al.}~\cite{sehwag2021ssd} combined contrastive learning  and the Mahalanobis distance for OOD detection.

There also exist a line of works~\cite{nalisnick2019detecting,huang2019out,serra2019input} which employed deep generative models on the pure unlabeled data. However, all of these methods require that the unlabeled data must be pure or OOD, which can hardly be met in realistic applications. Recently, some methods~\cite{chen2020semi,yu2020multi,guo2020safe}  considered the class distribution mismatch between labeled and unlabeled data, where the mismatched samples in the unlabeled data can be regarded as OOD samples. For example, Chen \textit{et al.}~\cite{chen2020semi} filtered out OOD samples in the unlabeled data with a confidence threshold and only utilised the remaining data for training.
Yu \textit{et al.}~\cite{yu2020multi} proposed a joint optimisation framework to classify identification samples and filter out OOD samples concurrently. Guo \textit{et al.} \cite{guo2020safe} employed bi-level optimization to weaken the weights of OOD samples. But these methods were developed for classifying identification samples and there were no OOD samples involved during the inference process. Yu~\textit{et al.}~\cite{yu2019unsupervised} attempted to utilise mixed unlabeled data for OOD detection, which encouraged two classifiers to maximally disagree on the mixed unlabeled data. However, since each unlabeled sample was treated equally, the model still required many labeled samples to distinguish between identification and OOD samples.

\section{Method}
In this section, we elaborate the major components of the proposed knowability-aware UniDA framework which sufficiently exploits the inter-sample affinity as stated in the introduction.

{\bf Notation} Assume that we have the labeled set of source samples $\mathcal{X}^s = { \left\{{{x^s_i}}\right\}}^{n^s}_{i=1} $ defined with the known space of the source label set $\mathcal{Y}^s$ and the unlabeled set of target samples $\mathcal{X}^t = \left\{{{x^t_i}}\right\}^{n^t}_{i=1}$ where $n^s$ and $n^t$ indicate the numbers of the source and the target samples, respectively.
Since the label spaces of the two domains are not aligned, we have the space of the target label set $\mathcal{Y}^t = \mathcal{Y}^{com} \cup \mathcal{Y}^{unk} $ with $\mathcal{Y}^{com} \subseteq \mathcal{Y}^{s}$. $\mathcal{Y}^{com}$ and $\mathcal{Y}^{unk}$ denote the spaces for the common label set {  which we called the known target label set} and the unknown label set respectively where $\mathcal{Y}^{unk} \cap \mathcal{Y}^{s} = \emptyset$. 
 {The known classes are the classes that exist in the source domain, where the learned model is expected to have knowledge of the labels for such classes. The known samples refer to the target samples that belong to the known classes. The unknown classes include the objective classes of some target samples that do not exist in the source domain, where the model does not learn the label information of such classes. The unknown samples refer to the target samples whose labels are unknown to the model.}
With the training samples from both domains, the goal of UniDA is to learn an optimal classifier $C^t : \mathcal{X}^t \rightarrow \mathcal{Y}^t$ which categorises a target sample into either the \textit{`unknown'} class or an object class belonging to $\mathcal{Y}^{com}$. 

\subsection{Overall Workflow}
\label{ow}

\begin{figure*}[t]
\centering
\includegraphics[width=1\linewidth]{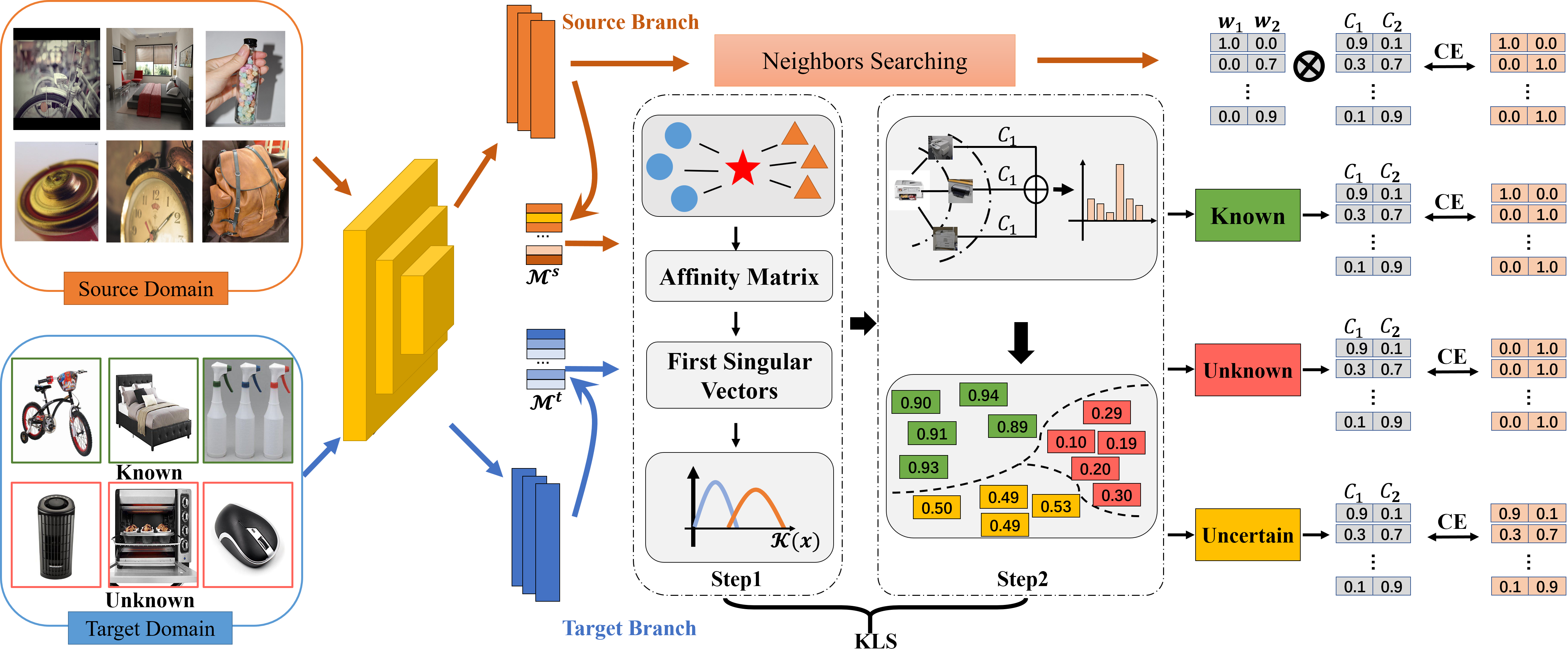}
\caption{The overall workflow of the proposed knowability-aware UniDA framework which exploits the inter-sample affinity. It leverages the knowability-guided detection of known/unknown samples and the label refinement based on neighborhood consistency to identify known samples and relabel them respectively. Both steps exploit the inter-sample affinity to obtain richer semantic information for every target sample. Finally, we use auxiliary losses to perform optimisation for our model to reduce the inter-sample affinity between the unknown and the known target samples.
}

\label{fig:frame}
\end{figure*}

As shown in Fig.~\ref{fig:frame}, we first extract a feature $ f_i $ from a sample $ x_i $ by the feature extractor $\mathcal{F}(\cdot\mid\phi)$ where $\cdot$ represents an input sample and $\phi$ denotes the set of trainable parameters of the feature extractor.
To perform an effective $k$-nearest neighbor search, we first build two memory banks $\mathcal{M}^s$ and $\mathcal{M}^t$ to store the features in the source and the target domains respectively:
\begin{equation}
    \mathcal{M}^s=[z^s_1,z^s_2, \cdots, z^s_{n^s}], \
    \mathcal{M}^t=[z^t_1,z^t_2, \cdots, z^t_{n^t}].
\end{equation}
which are updated by a momentum strategy:
 {\begin{equation}
    z_i^d = \alpha z_i^d + (1-\alpha) f_i^d,\quad f_i^d=F(x_i^d \mid \phi).
\label{update}
\end{equation}}
where $\alpha$ is the updating coefficient, $d \in \{s,t\}$. 

We then search the neighbors for each target sample from the two memory banks $\mathcal{M}^s$ and $\mathcal{M}^t$ to establish the affinity relationship between samples.  {Updating the memory banks is crucial for ensuring effective discrimination between features from different classes to find reliable neighbors by the $k$-nearest neighbors algorithm. The updating strategy of the memory bank in Eq. (\ref{update}) can progressively enhance the discrimination of features stored in the memory banks and reduce the intra-class variance between the given sample and its associated neighbors belonging to the same class from two domains. And the features with lower intra-class variance in the memory banks can effectively make the $k$-nearest neighbors algorithm more reliable. }

Next, we utilise the affinity relationship to perform the knowability-guided detection of known/unknown samples  and the label refinement based on neighborhood consistency. For the target samples, we categorise them into \textit{known}, \textit{unknown} and \textit{uncertain} classes based on the above two steps. We then design three losses, expressed as $\mathcal{L}_k$,
$\mathcal{L}_{unk}$
, and $\mathcal{L}_{unc}$
for the three classes of samples, which set desired restrictions on them respectively by exploiting the inter-sample affinities. Meanwhile, we establish an inter-sample affinity weight matrix $W_i$ for each sample in the source domain based on its neighbors, and then incorporate $W_i$ into the total loss $\mathcal{L}_s$.  Through minimising $\mathcal{L}_s$ during the training, the proposed method increases the inter-sample affinity within each class in the source domain whilst decreasing the inter-sample affinity between the samples of different classes in the source domain.
Finally, we employ one classifier $\mathcal{C(\cdot\mid\theta)}$ defined in Eq.~(\ref{classifier}) to classify all samples subject to the four losses:
\begin{equation}
\label{classifier}
\begin{aligned}
\mathcal{C}(\cdot \mid \theta): 
\boldsymbol{x} & \rightarrow\left[\begin{array}{cccc}
    \mathcal{C}_1^{(1)}(\cdot \mid \theta),  &... , & \mathcal{C}_1^{(Y)}(\cdot \mid \theta) \\ 
    \mathcal{C}_2^{(1)}(\cdot \mid \theta),  &... , &  \mathcal{C}_2^{(Y)}(\cdot \mid \theta)\\
\end{array}\right]^{T} 
\end{aligned}
\end{equation}
where the symbol $\theta$ denotes the set of parameters of the classifier implemented through a fully-connected layer. 
$ \mathcal{C}_1^{(j)}(\cdot\mid\theta)+\mathcal{C}_2^{(j)}(\cdot\mid\theta)=1 $, and  $\mathcal{C}_1^{(j)}$ and $\mathcal{C}_2^{(j)}$ represent the probabilities that a sample $x_i^t$ is accepted or rejected as a member of an object class with index $y$ 
in $\mathcal{Y}^{s}$ containing $Y$ object classes, respectively.  {Since $\mathcal{C}_1^{(j)}$ and $\mathcal{C}_2^{(j)}$ are output together, we use $\mathcal{C}_2^{(j)}$ to represent $1-\mathcal{C}_1^{(j)}$ for readability.} In the testing stage, for a target sample
$x^t_i$, we define the reject score of
$ x_i^t $ as the minimum  value of reject probabilities.
If $min_{j \in [1...Y]}(\mathcal{C}_2^{(j)}(x^t_i\mid\theta)) > 0.5 $,
we regard $x^t_i$ as an unknown target sample and otherwise a known target sample while the label $y_i = argmax_{j \in [1...Y]} (\mathcal{C}^{(j)}_1(x^t_i\mid\theta))$. 

\subsection{Knowability-Based Labeling Scheme}
In this section, we introduce the knowability-based labeling scheme (KLS) consisting of two steps which explore the label of a target sample based on the inter-sample affinity.

\subsubsection{Knowability-Guided Detection of  Known/Unknown Samples}
\label{step1}

\begin{figure}[t]
\centering
\includegraphics[width=0.8\linewidth]{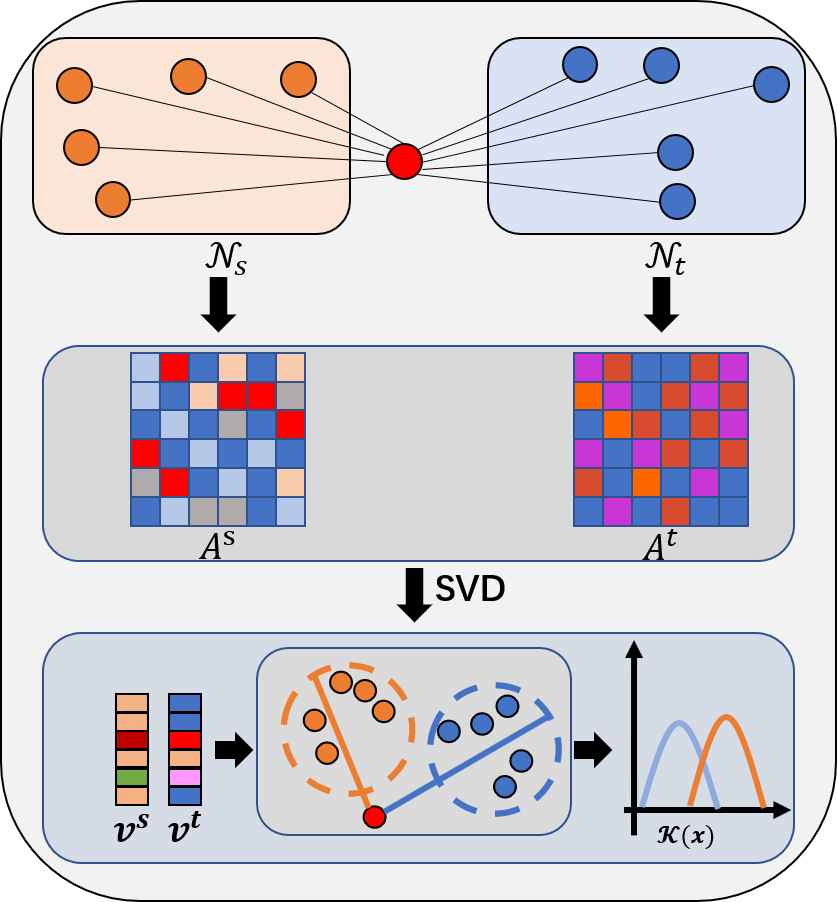}
\caption{ {Illustration of the computation of the knowability score. First, we search the neighbors of a target sample in both source and target domains. Then, we compute the affinity matrices of the neighbors in the source and the target domains, respectively. Next, we decompose each affinity matrix through SVD and obtain the first singular vectors of both matrices. Finally, we compute the knowability score defined as the cosine similarity of the two vectors.}}
\label{fig:setp1}
\end{figure}

To identify known and unknown samples, we explore the similarity of intrinsic structures of the neighborhood composed of source and target samples.  {With the assumption that the known target samples share similar semantics with the source samples, the distribution of the neighbors of a known sample from the target domain can be similar to that of a known sample from the source domain \cite{li2021DCC,wang2022cluster,zhao2021reducing,sharma2021instance}. }
To this end, we formulate the knowability-guided detection based on the consistency of intrinsic structures between {neighbors} searched from two domains.
In an effort to capture the intrinsic structure of neighbors, we propose to decompose the affinity matrices based on the $k$-nearest neighbors searched from both domains respectively to obtain the first singular vectors which robustly represent the intrinsic structures of the neighbors, as shown in Fig. \ref{fig:setp1}. In fact, the first singular vector has already been proven to be used to select representatives of the class \cite{zaeemzadeh2019iterative}. It is also used to obtain the degree of alignment between the representations and the eigenvector of affinity matrices of the representations for all classes, which uses the square of the inner product values between the representations and the first eigenvector to  detect credible and incredible instances \cite{kim2021fine}.

Specifically, given a target sample $x^t_i$, we first retrieve its {$k$} nearest neighbors from $\mathcal{M}^s$ and $\mathcal{M}^t$, denoted as $\mathcal{N}^s_i$ and $\mathcal{N}^t_i$, 
respectively:
\begin{equation}
\label{nsi}
     {\mathcal{N}^s_i=[z_{i0}^s,z_{i1}^s,\dots,z_{in}^s]^{T},\quad \mathcal{N}^t_i=[z_{i0}^t,z_{i1}^t,\dots,z_{in}^t]^{T},}
\end{equation}
 {where the sizes of $\mathcal{N}_i^s$ and $\mathcal{N}_i^t$ have to be equal. This may be a limitation in some applications.} Then, we compute the affinity matrices $A^s_i$ and $A^t_i$ for $\mathcal{N}^s_i$ and $\mathcal{N}^t_i$, respectively:
\begin{equation}
        \label{eq2}
            { A^s_i = \mathcal{N}^s_i{(\mathcal{N}^s_i)}^{T},\quad A^t_i = \mathcal{N}^t_i{(\mathcal{N}^t_i)}^{T}.}
\end{equation}
Next, we compute the first singular vectors of $A^s_{i}$ and $A^t_{i}$ via SVD decomposition as
\begin{equation}
A^s_i = {{U}}^s_i {{\Sigma}^s_i} {{V}}^s_i, \
A^t_i = {{U}}^t_i {{\Sigma}^t_i} {{V}}^t_i,
\label{sdec}
\end{equation}
 {where $\Sigma^s_i$ and $\Sigma^t_i$ are the decomposed diagonal matrices.} We obtain the first eigenvectors $v^s_i$, $v^t_i$ of ${{V}}^s_i$, ${{V}}^t_i$ corresponding to the largest eigenvalues.  {Note that it is unnecessary to sort $\mathcal{N}_i^s$ and $\mathcal{N}_i^t$ by similarity with $x_i^t$. We do not care about the sorting order of elements in $A^s_{i}$ and $A^t_{i}$ as we utilize the SVD method to decompose them and the decomposition is not affected by the order of the elements in the affinity matrix. If we change the sorting order of the two sets $\mathcal{N}_i^s$ and $\mathcal{N}_i^t$, it is equivalent to performing elementary matrix transformations for the matrices $A^s_{i}$ and $A^t_{i}$. Also, the singular vectors $v^s_i$ and $v^t_i$ are corresponding to the first singular values of $A^s_{i}$ and $A^t_{i}$, respectively, which are free of the orders of elements in $A^s_{i}$ and $A^t_{i}$.} 

The knowability score for the given samples $x^{t}_i$ can be produced by cosine similarity between $v^s_i$ and $v^t_i$:
\begin{equation}
  k(x^t_i)=\frac{{v^s_i}^{T} v^t_i}{\Vert v^s_i \Vert_2 \Vert v^t_i \Vert_2},
\label{cossim}
\end{equation}
We can observe that $k(x^t_i)$ represents the discrepancy of the semantic distributions between $\mathcal{N}^t_i$ and $\mathcal{N}^s_i$.  {Generally, when $k(x^t_i)$ becomes large, it means that the major directions of the feature distributions of $\mathcal{N}^s_i$ and $\mathcal{N}^t_i$ are very close. Otherwise, when $k(x^t_i)$ becomes small, $v_i^s$ is likely to be perpendicular to $v_i^t$, which means that the feature distributions of $\mathcal{N}^s_i$ is unrelated to that of $\mathcal{N}^t_i$.} Since the samples sharing the same semantic information (i.e. known target and source samples) are more likely to have similar distributions, $k(x^t_i)$ of known samples are larger than those of unknown samples which do not share any semantic information with source samples.
Thus, we divide these samples into known samples $\mathcal{D}_{known}$ and unknown samples $\mathcal{D}_{unknown}$ based on $k(x^t_i)$, respectively.

\subsubsection{Label Refinement Based on Neighborhood Consistency}
\label{step2}
Since the distribution of the known target samples can be less-discriminative compared to that of the source samples due to the domain bias, we propose a label refinement method based on the consistency of the predicted labels of the neighbors. In this stage, we further refine the labels of samples in $\mathcal{D}_{known}$, where we label the credible samples in $D_{known}$ and the samples from $D_{known}$ as the known samples.

In detail, for each sample $x^t_i$ in the target domain, 
we leverage the accepting probabilities of each sample from $\mathcal{N}^s_i$ produced by the classifier to compute the credibility score $c_i$:
\begin{equation}
\label{confidence}
    c_i = max_{j \in[1 ... Y]} \left(\frac{1}{ \mid \mathcal{N}_{i}^s \mid }{\sum}_{k \in \mathcal{N}_{i}} \mathcal{C}^{(j)}_1({z_k} \mid \theta)\right)
\end{equation}
where $\mathcal{N}^s_{i}$ denotes the set of indexes of the $k$-nearest neighbors in the source domain of the target sample $x^t_i$.

\begin{table}[]\Huge
    \centering
\setlength\tabcolsep{25mm}
\resizebox{0.45\textwidth}{!}{
\begin{tabular}{l}
\hline {\bf Algorithm 1} Algorithm of KLS\\
\hline {\bf Requirement:} $x_i^t$, $c_{\tau}$, $\mathcal{N}^s_i$,  $\mathcal{N}^t_i$\\
{\bf Step 1:}\\
Compute $A^s_i$, $A^t_i$\\
Decompose $A^s_i$ and $A^t_i$ by Eq. (\ref{sdec})\\
Obtain $v^s_i$,$v^t_i$\\
Compute the knowability-score $k(x^t_i)$ by Eq. (\ref{cossim})\\
{\bf If} $k(x^t_i)<k_{\tau}$ do\\
$\quad $ Append $x^t_i$ to $\mathcal{D}_{unknown}$\\
{\bf Else} do \\
$\quad $ Append $x^t_i$ to $\mathcal{D}_{known}$\\
{\bf Step 2:}\\
$\quad $ Compute $c_{\tau}$ by Eq.(\ref{ctau})\\
{\bf If} $x^t_i\in \mathcal{D}_{known}$ do\\
$\quad $ Compute $c_i$ by Eq. (\ref{confidence})\\
$\quad $ {\bf If} $c_i>c_{\tau}$ do\\
$\qquad $ Obtain the pseudo label $\hat{y}^t_i$\\
$\qquad $ Label $x_i^t$ as $\hat{y}^t_i$\\
$\quad $ {\bf Elif} $c_i<0.8c_{\tau}$ do\\\
$\qquad $ Label $x_i^t$ as \textit{`Unknown'}\\
$\quad $ {\bf Else} do\\
$\qquad $ Label $x_i^t$ as \textit{`Uncertain'}\\
{\bf Elif} $x^t_i\in \mathcal{D}_{unknown}$ do\\
$\quad $ Label $x_i^t$ as \textit{`Unknown'}\\
end\\
\hline
\end{tabular}}
\end{table}

The lower $c_i$ indicates that the predicted label of the target sample is highly dissimilar to any known class, suggesting that the target sample may lie near the decision boundary of the model. We identify such samples as unknown samples. In contrast, a target sample with a higher $c_i$ is likely to be far away from the decision boundary and can derive a more reliable pseudo label from its neighbors. Formally, if $c_i< 0.8c_{\tau}$, we regard $x^t_i$ as an unknown sample. Note that the threshold $c_{\tau}$ is produced  automatically and $0.8$ is chosen empirically. Then, if $c_i > c_{\tau}$, $x^t_i$ is recognised as a known sample. If $0.8 c_{\tau} < c_i < c_{\tau}$, $x^t_i$ is regarded as an uncertain sample (sensitivity of the scale coefficient for $c_{\tau}$ can be seen in Sec. \ref{ablation}).

Distinguishing the unknown samples from the known ones in the target domain is obviously affected by the choice of the threshold $c_\tau$. However, varying sizes and categories of different datasets lead to the change of the optimal threshold. To avoid setting the threshold manually for each dataset, we introduce an auto-thresholding scheme. Notably, the threshold $c_{\tau}$ is calculated as the mean of the maximum values for the accepting probabilities $\mathcal{C}_1(x_i^s\mid \theta)$ of source samples in the mini-batch $\mathcal{B}$:
\begin{equation}
\label{ctau}
    c_{\tau}=\frac{1}{\mid \mathcal{B} \mid} \sum_{i=1}^{\mid \mathcal{B} \mid} \max _{j \in[1 . . Y]}\left(\mathcal{C}^{(j)}_1\left(x^s_i\mid \theta\right)\right).
\end{equation}

\begin{figure}[t]
  \centering
   \includegraphics[width=1\linewidth]{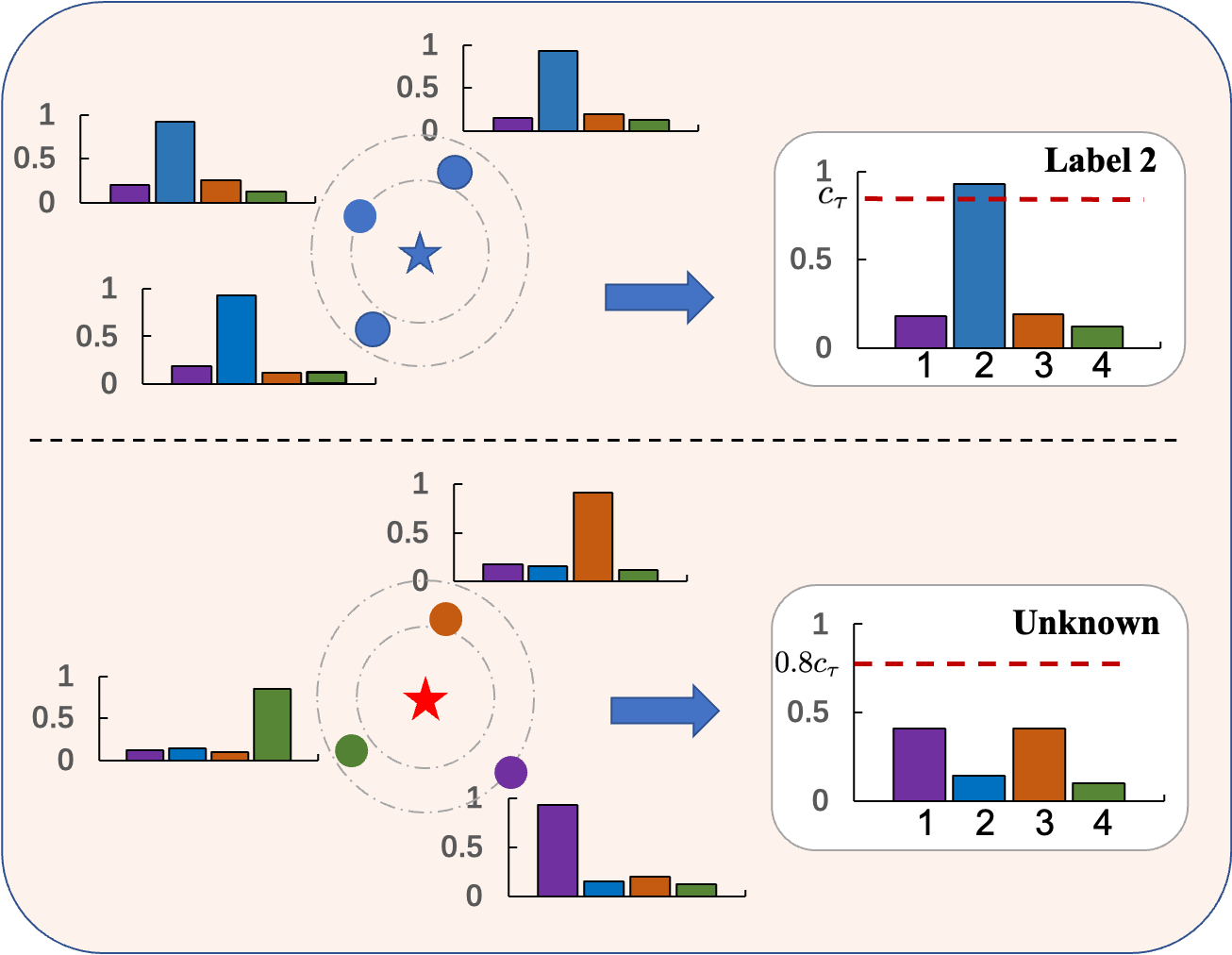}
   \caption{Overview of the label refinement. We find the $k$-nearest neighbors from the source domain for each target sample. $c_i$ is computed as the maximum value of the average accepting probabilities of the neighbors of each target sample. 
   }
   \label{fig:setp2}
\end{figure}

This scheme avoids setting different thresholds for different datasets manually. This step is also illustrated in Fig. \ref{fig:setp2} and the full algorithm of KLS is elaborated in Algorithm 1.

\subsection{Training Objectives}
\subsubsection{Target Domain Losses}

Once we derive the known and the unknown samples from the above two steps, we propose the auxiliary losses to reduce the inter-sample affinity between~the unknown and the~ known samples ~ and increase that within a known class. 
Specifically, for an unknown sample, we hope to push the samples of all known classes away from it for reducing the inter-sample affinity between the unknown and the known samples. Thus we design the target-domain loss for the unknown samples, $\mathcal{L}_{unk}$, which minimizes the entropy of the reject probabilities for all classes:
\begin{equation}
\label{lunk}
    \mathcal{L}_{unk}(x^t_i) = -\frac{
    1}{Y}\sum_{j=1}^{Y}\mathcal{C}_2^{(j)}(x^t_i \mid \theta)log\left( \mathcal{C}_2^{(j)}(x^t_i \mid \theta) \right).
\end{equation}

For the known samples in the target domain, we define the pseudo label of $x^t_i$ as: 
\begin{equation} 
\label{pseudo}
\hat{y}^t_i = {argmax}_{j \in[1 ... Y]}({\sum}_{k \in \mathcal{N}_{i}} \mathcal{C}^{(j)}_1(x^s_k \mid \theta))
\end{equation}
where $argmax(\cdot)$ denotes the index of the biggest value in a vector. Since the discrepancies exist between the source and the target samples belonging to the same object class due to the domain gap, the inter-sample affinity between them cannot be as high as that between the source samples belonging to the same object class. Thus, to increase the inter-sample affinity within a known class in the target domain, we increase the inter-sample affinity between the known samples in the target domain and the corresponding samples with the pseudo label $\hat{y}_i^t$ in the source domain. This is achieved by designing the target-domain loss $\mathcal{L}_k$ which minimizes the entropy of the accepting probability of class $\hat{y}^t_i$:
\begin{equation}
\label{lk}
    \mathcal{L}_{k}(x^t_i) = - \mathcal{C}_1^{(\hat{y}_i^t)}(x^t_i \mid \theta)log\left( \mathcal{C}_1^{(\hat{y}_i^t)}(x^t_i \mid \theta) \right).
\end{equation}

 {Moreover, it is difficult to distinguish uncertain samples as known or unknown ones. Therefore, we apply the self-supervised learning to minimize the sum of the average entropy of $\mathcal{C}_1^{(j)}$ and $\mathcal{C}_2^{(j)}$. Since $\mathcal{C}_1^{(j)}+ \mathcal{C}_2^{(j)}=1$ for any given class, by minimizing the entropy, the uncertain samples supposed to be known will have an increase in the confidence of belonging to one source class, while the uncertain samples supposed to be unknown will have an increase in the reject scores of each class. As such, the uncertain samples can be distinguished more reliably.}
We leverage a loss $\mathcal{L}_{unc}$ to minimize the average entropy of all classifiers to keep the inter-sample affinities low in every known classes: 

\begin{equation}
\mathcal{L}_{unc}(x^t_i) = \hspace{-1mm} \frac{-1}{2Y}\hspace{-1mm} \sum_{ k=1,2}\hspace{-0.5mm}\sum_{j=1}^{Y} \mathcal{C}_k^{(j)}(x^t_i \mid \theta)log \hspace{-1mm}\left( \mathcal{C}_k^{(j)}(x^t_i \mid \theta) \right)\hspace{-1mm}.\hspace{-1mm}
\end{equation}

The overall algorithm of our method is elaborated in Algorithm 2.

\begin{table}[]\Huge
    \centering
\setlength\tabcolsep{8mm}
\resizebox{0.45\textwidth}{!}{
\begin{tabular}{l}
\hline {\bf Algorithm 2} Full algorithm of our method\\
\hline {\bf Requirement}:  ($\mathcal{X}^{s}$, $\mathcal{Y}^{s}$), $\mathcal{X}^{t}$  \\
{\bf while} step $<$ max step do\\
$ \quad$Sample batch $\mathcal{B}^s$ from ($\mathcal{X}^{s}$, $\mathcal{Y}^{s}$) and batch $\mathcal{B}^t$ from $\mathcal{X}^{t}$ \\
$ \quad$Extract features from each of $\mathcal{B}^s$and $\mathcal{B}^t$ \\
$ \quad${\bf If} step $== 0$ do \\
$ \qquad$ {Initialize} $\mathcal{M}^t$, $\mathcal{M}^s$ \\
$ \quad${\bf Else} do \\
$ \qquad$ Update $\mathcal{M}^t$, $\mathcal{M}^s$ \\
$ \quad${\bf for} $x^{s}_i \in \mathcal{B}^s$ and $x^{t}_i \in \mathcal{B}^t$ do \\
$ \qquad$Compute  $W_i$ for $x^{s}_i$   \\
$ \qquad$ {Compute the source domain loss $\mathcal{L}_s$ } \\
$ \qquad$Label $x^t_i$ by KLS \\
$ \qquad${\bf If} $x^t_i$ has label $\hat{y}_i^s$ do \\
$ \qquad\quad$Compute $\mathcal{L}_k$ \\
$ \qquad${\bf Elif} $x^t_i$ has label \textit{`unknown'} do \\
$ \qquad\quad$Compute $\mathcal{L}_{unk}$  \\
$ \qquad${\bf Else} do \\
$ \qquad\quad$Compute $\mathcal{L}_{unc}$ \\

$ \qquad$Compute the overall loss $\mathcal{L}_{all}$\\
$ \quad$Update the model\\
end\\
\hline
\end{tabular}}
\end{table}

\subsubsection{Source Domain Loss based on Inter-sample Affinity}

For a sample $x_i^s$ in the source domain with label $y^s_i$, to deliver a reliable classification, we should increase the inter-sample affinity within class $y^s_i$ and reduce that between class $y^s_i$ and other classes in the source domain. Thus, we propose the inter-sample affinity weight matrix 
$W_i = [w_1, w_2]^{T}$ for $x_i^s$ where $w_1, w_2 \in \mathbb{{R}}^Y$ represent the weights associated with the classes which require to increase or decrease the inter-sample affinity, respectively. In detail,  
$w_1 = \left (\mathbf{1}(j=y^s_i) \right)_{j=1}^Y$ is the one-hot vector of class $y^s_i$. And $w_2 = \left(w_2^{(j)}\right)^Y_{j=1}$  is computed based on the inter-sample affinities between $x^s_i$ and the samples from other source classes by retrieving the $k$-nearest neighbors of $x^s_i$ from the samples with the labels different from $ y^s_i$ in the source domain, expressed as:
\begin{equation}
\label{weight}
    w_2^{(j)} = norm(\frac{
    \mid \mathcal{N}_i^{(j)} \mid }{\mid \mathcal{N}_i \mid}*\frac{\mathcal{C}_1^{(j)}(x^s_i \mid \theta) }{\sum_{k \neq y^s_i} \mathcal{C}_1^{(k)}(x^s_i \mid \theta)^2})
\end{equation}
 {where $norm$ denotes the L1-normalisation and $*$ is the multiplication.} $ \mid \mathcal{N}_i^{(j)} \mid$ and $\mid \mathcal{N}_i \mid$ represent the number of the neighbors belonging to the label $y_j^s$ and the total number of the retrieving neighbors of $x^s_i$ respectively and note that $w_2^{y_i^s}$ is set to $0$.
According to the Eq.~(\ref{weight}), the larger values in $w_2$  means that the samples in class $j$ are closer to $x^s_i$. Then, we compute the source-domain loss $\mathcal{L}_{s}(x^s_i)$ based on the weighted inter-sample affinity:
\begin{equation}
\label{ls}
    \mathcal{L}_s(x^s_i) = - \log < W_i,\mathcal{C}(x_i^s \mid \theta)  > 
\end{equation}
where $\langle\cdot, \cdot\rangle$ is the dot product operator. 

\subsection{Overall Loss for Both Domains}

Overall, we train the classifier $\mathcal{C}(\cdot \mid \theta)$ and the feature extractor $\mathcal{F}(\cdot \mid \phi)$ with four losses and a hyper-parameter $\lambda$. The overall loss is expressed as:
\begin{equation}
  \mathcal{L}_{all} = \mathcal{L}_s + \lambda (\mathcal{L}_{unk} + \mathcal{L}_{k} + \mathcal{L}_{unc}).
\end{equation}
It is worth mentioning that differing from many existing UniDA methods~\cite{li2021DCC,bucci2020effectiveness,fu2020learning,saito2020dance}, there is only one hyper-parameter in our method.

\section{Experimental Results}
We do experiments on several  benchmarks, such as Office-31 (Saenko et al. \cite{saenko2010adapting}), OfficeHome (Peng et al. \cite{peng2019moment}), VisDA (Peng et al. \cite{peng2017visda}) and DomainNet (Venkateswara et al. \cite{venkateswara2017deep}).
In this section, we first introduce our experimental setups, including datasets, evaluation protocols and training details. Then, we compare our method with a set of the state-of-the-art (SOTA) UniDA methods. We also conduct extensive ablation studies to demonstrate the effectiveness of each component of the proposed method. All  experiments were implemented on one RTX2080Ti 11GB GPU with PyTorch 1.7.1 \cite{paszke2019pytorch}.

\subsection{Experimental Setups}
\subsubsection{Datasets and Evaluation Protocols}
\label{datasetde}
We conduct experiments on four datasets. Office-31 \cite{saenko2010adapting} consists of $4,652$ images from three domains: DSLR (D), Amazon (A), and Webcam (W). OfficeHome \cite{peng2019moment} is a more challenging dataset, which consists of $15,500$ images from $65$ categories. It is made up of $4$ domains: Artistic images (Ar), Clip-Art images (CI), Product images (Pr), and Real-World images (Rw). VisDA \cite{peng2017visda} is a large-scale dataset, where the source domain contains $15,000$ synthetic images and the target domain consists of $5,000$ images from the real world. DomainNet \cite{venkateswara2017deep} is a larger DA dataset containing around $0.6$ million images.

In this paper, we use the H-score in line with recent UniDA methods \cite{fu2020learning,li2021DCC,Saito_2021_ICCV}. H-score, proposed by Fu \textit{et al.} \cite{fu2020learning}, is the harmonic mean of the accuracy on the common classes $a_{com}$ and the accuracy on the unknown class $a_{unk}$:
\begin{equation}
    h = \frac{2a_{com}\cdot a_{unk}}{a_{com} + a_{unk}}.
    \label{h-score}
\end{equation}

\subsubsection{Training Details} We employ the ResNet-50 \cite{he2016deep} backbone pretrained on ImageNet \cite{deng2009imagenet} and optimise the model using Nesterov momentum SGD with momentum of $0.9$ and weight decay of $5 \times 10^{-4}$ . The batch size is set to $36$ for all datasets. The initial learning rate is set as $0.01$ for the new layers and $0.001$ for the backbone layers. The learning rate is decayed with the inverse learning rate decay scheduling. The updating coefficient $\alpha$ is set as $0.9$. The number of neighbors retrieved is set differently for different datasets.
For Office-31 ($4,652$ images in $31$ categories) and OfficeHome ($15,500$ images in $65$ categories), the numbers of retrieved neighbors (i.e., $\mid \mathcal{N}^s_i \mid$, $\mid \mathcal{N}^t_i \mid$ and $\mid \mathcal{N}_i \mid$) are all set to $10$. For VisDA ($20,000$ images in total) and DomainNet ($0.6$ million images), we set them to $100$, respectively. $k_\tau$ is set to $0.5$ for all datasets. We set $\lambda$ to  $0.1$ for all datasets.
\begin{table}[h]
    \caption{Results on Office-31 with UniDA setting (H-score).}
   \centering
   \setlength{\tabcolsep}{1.1mm}
   {
   \begin{tabular}{l|cccccc|c}
    ine \multirow{2}*{ Method } & \multicolumn{6}{c|}{ Office-31 $(10 / 10 / 11)$} & \multicolumn{1}{l}{} \\
    & A2D & A2W & D2A & D2W & W2D & W2A & Avg \\
    \hline UAN \cite{you2019universal} & $59.7$ & $58.6$ & $60.1$ & $70.6$ & $71.4$ & $60.3$ & $63.5$ \\
    CMU \cite{fu2020learning} & $68.1$ & $67.3$ & $71.4$ & $79.3$ & $80.4$ & $72.2$ & $73.1$ \\
    DANCE \cite{saito2020dance} & $78.6$ & $71.5$ & $79.9$ & $91.4$ & $87.9$ & $72.2$ & $80.3$ \\
    DCC \cite{li2021DCC} & $8 8 . 5$ & $78.5$ & $70.2$ & $79.3$ & $88.6$ & $75.9$ & $80.2$ \\
    ROS \cite{bucci2020effectiveness} & $71.4$ & $71.3$ & $81.0$ & $94.6$ & $9 5 . 3$ & $79.2$ & $82.1$ \\
    USFDA \cite{kundu2020universal} & $85.5$ & $7 9 . 8$ & $\mathbf{83.2}$ & $90.6$ & $88.7$ & $81.2$ & $84.8$ \\
    OVANet \cite{Saito_2021_ICCV} & $85.8$ & $79.4$ & $80.1$ & ${95.4}$ & $94.3$ & $8 4 . 0$ & $8 6 . 5$ \\[0.5pt]
    \hline
    Ours & $\mathbf{87.4}$ & $\mathbf{82.5}$ & $80.6$ & $\mathbf{96.1}$ & $\mathbf{98.3}$ & $\mathbf{84.9}$ & $\mathbf{88.5}$ \\
   \hline
    \end{tabular}}
    \label{tab:office-31}
  \end{table}

 \begin{table*}[t]
\centering
    \caption{Results on OfficeHome with UniDA setting (H-score).}
{
\begin{tabular}{l|cccccccccccc|c}

\hline \multirow{2}{*}{ Method } & \multicolumn{12}{c|}{OfficeHome (10/5/50) } & \multicolumn{1}{l}{} \\[0.5pt]
& A2C & A2P & A2R & C2A & C2P & C2R & P2A & P2C & P2R & R2A & R2C & R2P & Avg \\[0.5pt]
\hline OSBP\cite{saito2018open} & $39.6$ & $45.1$ & $46.2$ & $45.7$ & $45.2$ & $46.8$ & $45.3$ & $40.5$ & $45.8$ & $45.1$ & $41.6$ & $46.9$ & $44.5$ \\[0.5pt]
UAN\cite{you2019universal} & $51.6$ & $51.7$ & $54.3$ & $61.7$ & $57.6$ & $61.9$ & $50.4$ & $47.6$ & $61.5$ & $62.9$ & $52.6$ & $65.2$ & $56.6$ \\[0.5pt]
CMU\cite{fu2020learning} & $56.0$ & $56.9$ & $59.1$ & $66.9$ & $64.2$ & $67.8$ & $54.7$ & $51.0$ & $66.3$ & $68.2$ & $57.8$ & $69.7$ & $61.6$ \\[0.5pt]
OVANet\cite{Saito_2021_ICCV} & $6 2 . 8$ & $75.6$ & $78.6$ & ${70.7}$ & $68.8$ & $75.0$ & $71.3$ & $58.6$ & $80.5$ & $76.1$ & $64.1$ & $78.9$ & $71.8$ \\[0.5pt]
\hline 
 Ours  & $\mathbf{64.3}$ & $\mathbf{80.4}$ & $\mathbf{86.1}$ & $\mathbf{72.0}$ & $\mathbf{71.1}$ & $\mathbf{77.8}$ & $\mathbf{71.5}$ & $\mathbf{61.7}$ & $\mathbf{8 3 . 8}$ & $\mathbf{79 . 1}$ & $\mathbf{6 4.8}$ & $\mathbf{82.4}$ & $\mathbf{7 4 . 6}$ \\[0.5pt]
\hline
\end{tabular}}
    \label{tab:officehome}
\end{table*}

\begin{table*}[t]

   \centering
       \caption{Results on DomainNet and VisDA with UniDA setting (H-score).}
    \setlength{\tabcolsep}{3.2mm}
    {
    \begin{tabular}{l|cccccc|c||c}
\hline \multirow{2}{*}{ Method } & \multicolumn{6}{c|}{DomainNet $(150 / 50 / 145)$} & & VisDA \\[0.4pt]
& $\mathrm{P} 2 \mathrm{R}$ & $\mathrm{R} 2 \mathrm{P}$ & $\mathrm{P} 2 \mathrm{S}$ & $\mathrm{S} 2 \mathrm{P}$ & $\mathrm{R} 2 \mathrm{S}$ & $\mathrm{S} 2 \mathrm{R}$ & $\mathrm{Avg}$ & $(6 / 3 / 3)$ \\
\hline DANCE \cite{saito2020dance} & $21.0$ & $47.3$ & $37.0$ & $27.7$ & $\mathbf{4 6 . 7}$ & $21.0$ & $33.5$ & $4.4$  \\
UAN \cite{you2019universal} & $41.9$ & $43.6$ & $39.1$ & $38.9$ & $38.7$ & $43.7$ & $41.0$ & $30.5$  \\
CMU \cite{fu2020learning} & $50.8$ & $5 2 . 2$ & $45.1$ & $44.8$ & $45.6$ & $51.0$ & $48.3$ & $34.6$\\
DCC \cite{li2021DCC} &$5 6 . 9$ & $50.3$ & $43.7$ & $44.9$ & $43.3$ & $56.2$ & $49.2$ & $43.0$  \\
OVANet \cite{Saito_2021_ICCV} & $56.0$ & $51.7$ & $4 7 . 1$ & $4 7 . 4$ & $4 4 . 9$ & $5 7 . 2$ & $5 0 . 7$ & $5 3 . 1$  \\
\hline 
 Ours & $\mathbf{59.1}$ & $\mathbf{52.4}$ & $\mathbf{4 7 . 5}$ & $\mathbf{4 8 . 1}$ & $45.1$ & $\mathbf{58.6}$ & $\mathbf{5 1 . 8}$ & $\mathbf{5 4 . 7}$  \\[0.4pt]
\hline
    \end{tabular}}
    \label{tab:dnet}
\end{table*}

\begin{figure*}[t]

\subfigure[Ours]{
\begin{minipage}[t]{0.33\linewidth}
\centering
\includegraphics[width=2.2in]{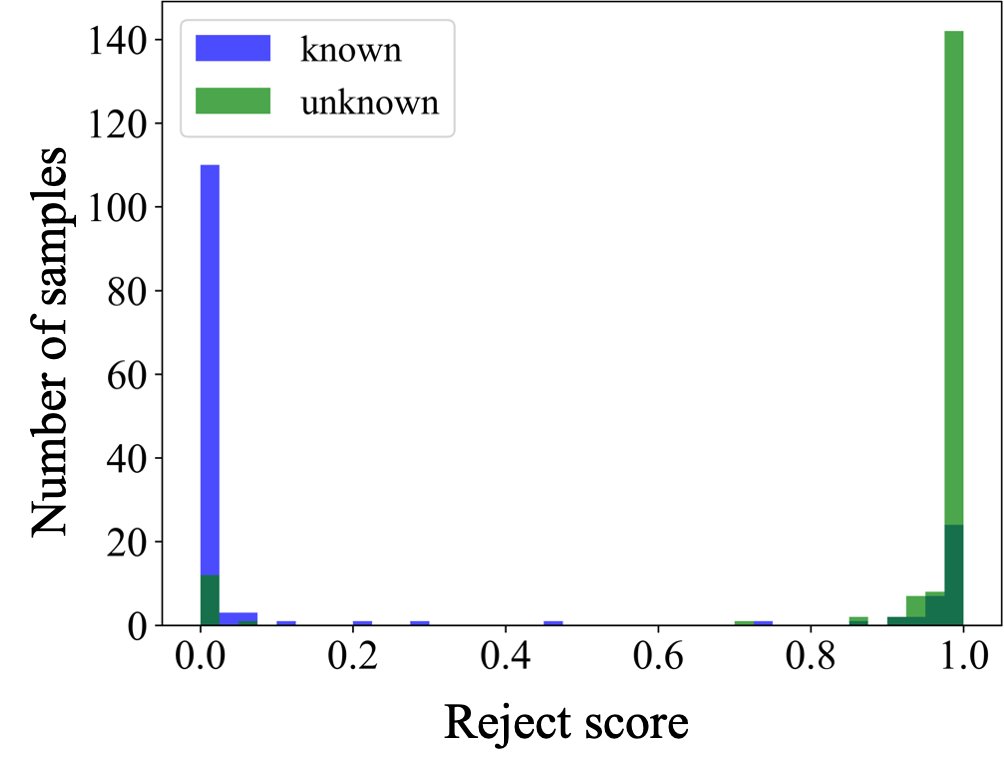}
\end{minipage}%
}%
\hspace{-1mm}
\subfigure[Source Only]{
\begin{minipage}[t]{0.33\linewidth}
\centering
\includegraphics[width=2.2in]{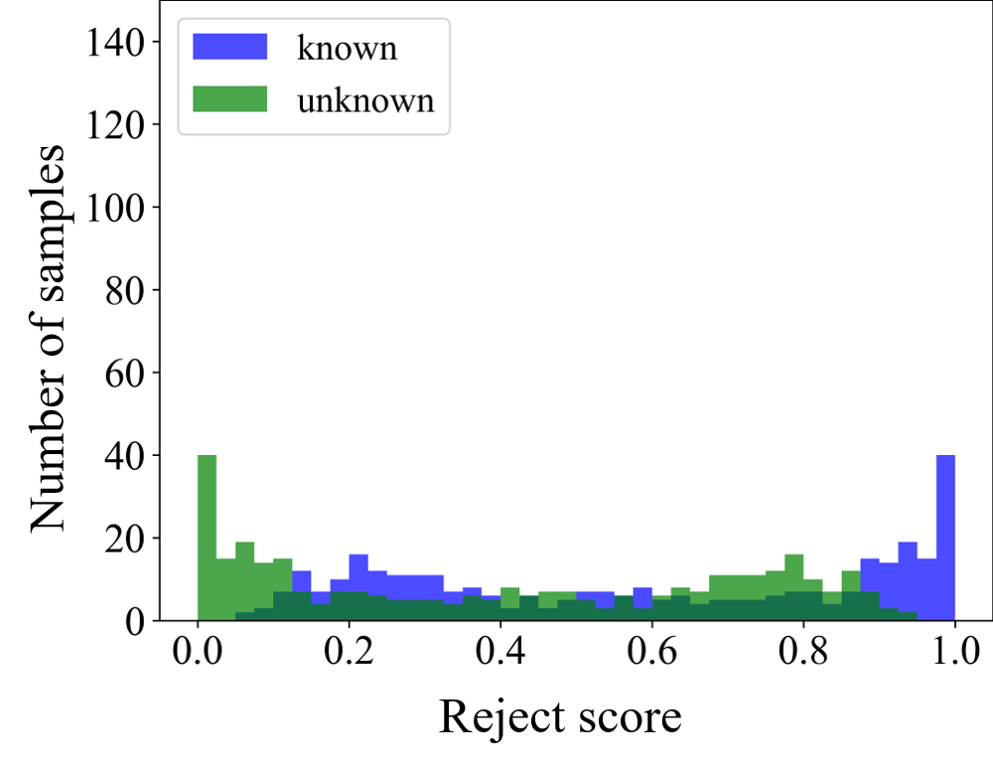}
\end{minipage}%
}%
\hspace{-1mm}
\subfigure[OVANet]{
\begin{minipage}[t]{0.33\linewidth}
\centering
\includegraphics[width=2.2in]{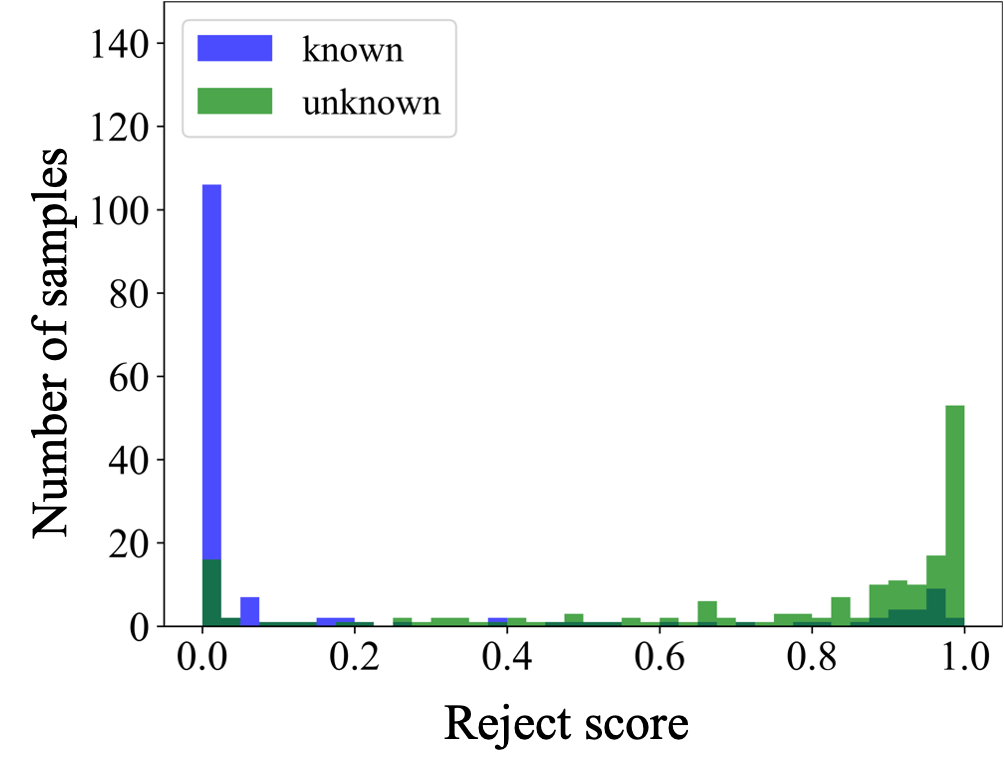}
\end{minipage}%
}%

\centering
\caption{Comparison on the distribution of reject scores. The three plots of histograms show the reject scores at the last epoch produced by the full version of our method, the model trained only on source domain, and the model trained 
 on OVANet~\cite{Saito_2021_ICCV} in Office-31(A2D) respectively. Each area in dark green indicates that there is an overlap between the green and the blue bars.}
\label{ed}
\end{figure*}

\subsection{Comparison with the SOTA Methods}
\subsubsection{{Baselines}}  
We compare our method with several SOTA methods under the same settings on the four
datasets in Sec.~\ref{datasetde}, such as UAN~\cite{you2019universal}, CMU~\cite{fu2020learning} and DCC~\cite{li2021DCC}.
We aim to show that the knowability-based labeling scheme (KLS) is effective for UniDA, which employed a classifier to produce the confidence of each sample to determine whether it belongs to the unknown class or not. Also, we compare our method with OVANet~\cite{Saito_2021_ICCV} and DANCE~\cite{saito2020dance} to show that it is important to reduce the inter-sample affinity  between the unknown and the known samples.

\subsubsection{Results in Main Datasets}

Tables \ref{tab:office-31} and \ref{tab:officehome}  list the results on Office-31 and OfficeHome, respectively. On Office-31, our method outperforms the SOTA methods by  {$2.0\%$} in terms of the H-score on average. For the more challenging dataset OfficeHome which contains much more private classes than common classes, our method also made a significant improvement of $2.8\%$ in terms of the H-score.
Our method also achieves the SOTA performance on both VisDA AND DomainNet as shown in Table \ref{tab:dnet}. Overall, according to the results of quantitative comparisons, our method achieves the SOTA performance in every dataset and most sub-tasks, which demonstrates 
the effectiveness of the main idea of our method that reduces the inter-sample affinity between the unknown and the known samples.
\vspace{-5pt}
\subsection{Ablation Studies}
\label{ablation}
In this section, we provide specific analysis on several important issues and ablated studies to understand the behaviour of our method.

\begin{table*}[t]
    \centering
\caption{ {Evaluation of models trained on Office-31 using new testing samples from the subsets `Art' and `Clipart' of OfficeHome.}}
   \setlength{\tabcolsep}{4.2mm}
   {
   \begin{tabular}{l|l|cccccc|c}
   \hline
    Testing  &{} & \multicolumn{6}{c|}{Training Sub-tasks on Office-31 $(10 / 10 / 11)$} & \multicolumn{1}{l}{} \\
    {{Datasets}} &\multirow{-2}{*}{Method}&A2D & A2W & D2A & D2W & W2D & W2A & Avg \\
    \hline
     & Ours & $72.2$ & $74.6$ & $100$ & $100$ & $99.1$ & $99.1$ & $96.9$\\
     
    \multirow{-2}{*}{Art } &OVANet \cite{Saito_2021_ICCV}& $61.6$ & $71.2$ & $99.1$ & $98.3$ & $100$ & $99.1$ & $88.2$\\
     
    \hline
    & Ours &  $67.4$ & $72.0$ & $100$ & $99.6$ & $98.7$ & $100$ & $89.6$\\
     \multirow{-2}{*}{Clipart } &{OVANet \cite{Saito_2021_ICCV}}&  $69.1$ & $70.4$ & $99.6$ & $100$ & $98.7$ & $99.1$ & $89.4$\\
    \hline
    \end{tabular}}
    \label{newdata}
\end{table*}

\begin{table*}[t]
    \centering
\caption{ {Results produced with different values of $k$ on Office-31.}}
   \setlength{\tabcolsep}{3.2mm}
   {
   \begin{tabular}{c|ccccccccccccc}
   \hline
    { $k$ }
    & $5$ & $7$ & $9$ & $10$ & $11$ & $13$ &$15$ &$20$ &$30$ &$40$ &$50$ &$70$ &$90$  \\
    \hline 
     A$2$D& $87.0$ & $87.2$ & $87.6$ & $87.4$ & $87.4$ & $87.9$ &$87.8$ &$\mathbf{88.2}$ &$87.2$ &$86.4$ &$86.0$ &$85.2$ &$84.0$ \\
     D$2$A& $80.7$ & $80.8$ & $80.5$ & $80.6$ & $80.2$& $\mathbf{80.8}$ &$80.1$ &$79.5$ &$78.9$ &$78.0$ &$77.2$ &$75.0$ &$72.5$ \\
     \hline
     D$2$W& $96.2$ & $\mathbf{96.6}$ & $96.3$ & $96.1$ & $96.3$ & $96.5$ &$96.5$ &$95.8$ &$95.3$ &$94.2$ &$93.0$ &$91.2$ &$89.9$ \\
     W$2$D& $98.1$ & $98.0$ & $98.2$ & $\mathbf{98.3}$ & $98.3$ & $98.1$ &$97.6$ &$97.3$ &$96.5$ &$96.1$ &$95.2$ &$93.6$ &$92.8$ \\   
    \hline
    \end{tabular}}
    \label{tab:knn}
\end{table*}

\vspace{8mm}

{\bf Quantitative Comparison on the Distribution of Reject Scores.}  To show the improvement on the distribution of reject scores which is the confidence of classifying the unknown samples as introduced in Sec.~\ref{ow}, we conducted experiments on Office-31(A2D). First, we plot the distributions of the reject scores of all sample in the target domain at the final epoch in Fig.~\ref{ed} {\bf (a)}. Then, we compare the plot to that trained on the source domain only in Fig.~\ref{ed} {\bf(b)}. We can observe that the full version of our method better distinguishes the known samples from the unknown ones. Furthermore, in Fig.~\ref{ed} {\bf (c)}, we show the corresponding plot produced by OVANet~\cite{Saito_2021_ICCV} for comparison. Noticeably, our method performs better than OVANet~\cite{Saito_2021_ICCV} in terms of distinguishing the known samples from the unknown ones.  {However, it can be seen from Fig.~\ref{ed} that negative transition also occurs, corresponding to the overlapping regions between the blue and the green bars. Such overlaps indicate that known samples are misclassified as unknown samples, or vice versa. Domain gap is the primary reason for the observed negative transition, which hinders the accurate classification of known and unknown samples.}

 {\textbf{Ability of Detecting Completely New Unknown Samples.} To further show our model's ability of detecting completely new unknown samples not included in the training dataset, we conduct experiments using completely new testing datasets and show the results in Table \ref{newdata}. Specifically, the model was trained on Office-31 and tested on the subsets `Art' and `Clipart' of OfficeHome. Both subsets comprise samples that do not belong to any known classes. It can be seen that our method performs well in detecting completely new unknown samples, showcasing superior performance compared to the recent baseline OVANet \cite{Saito_2021_ICCV}.}

\begin{figure}[t]
\centering

\includegraphics[width=0.9\linewidth]{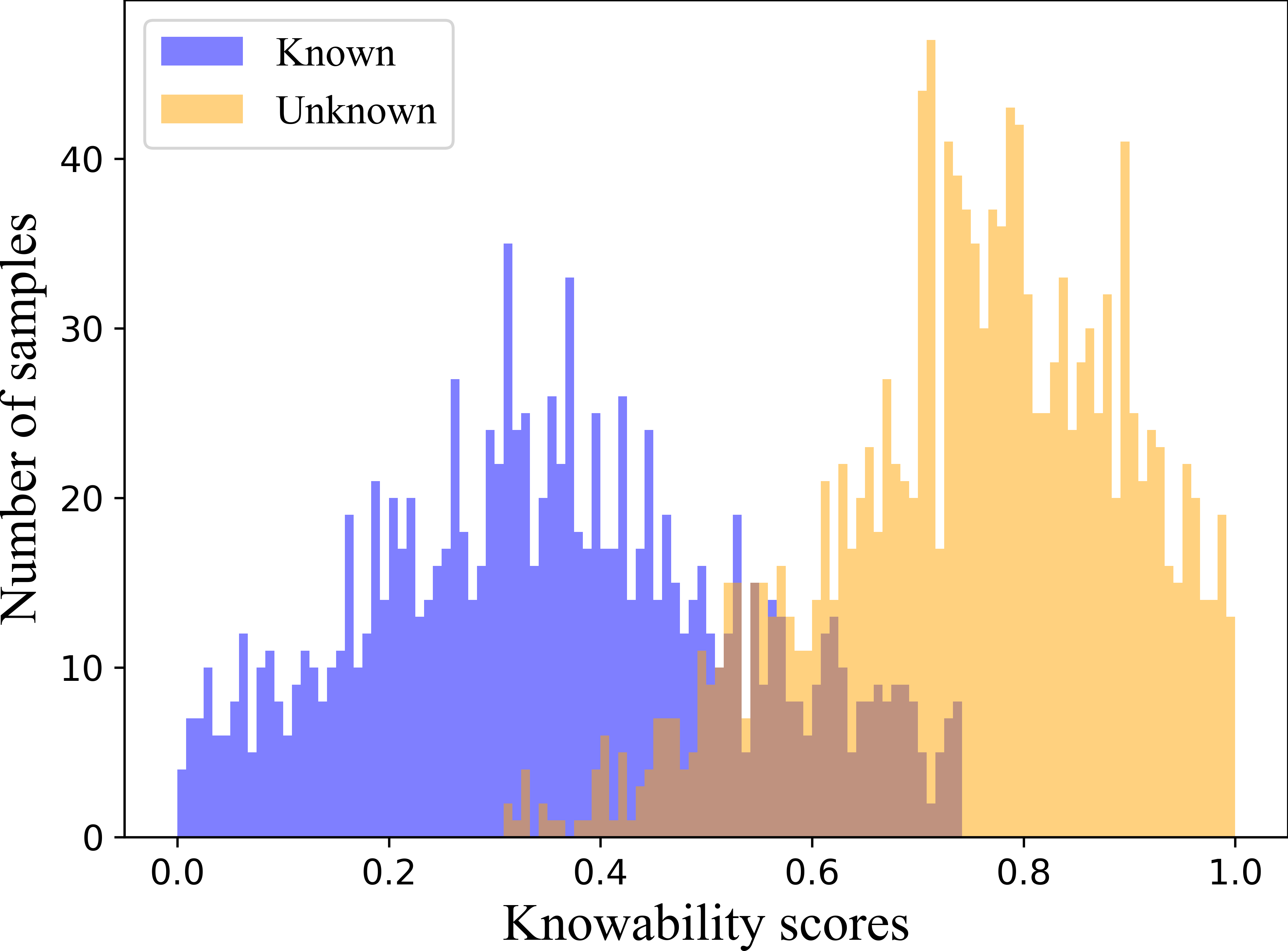}

\caption{ The distribution of the knowability score.}
\label{jj}
\end{figure}

\begin{figure}[t]
\centering

\includegraphics[width=0.9\linewidth]{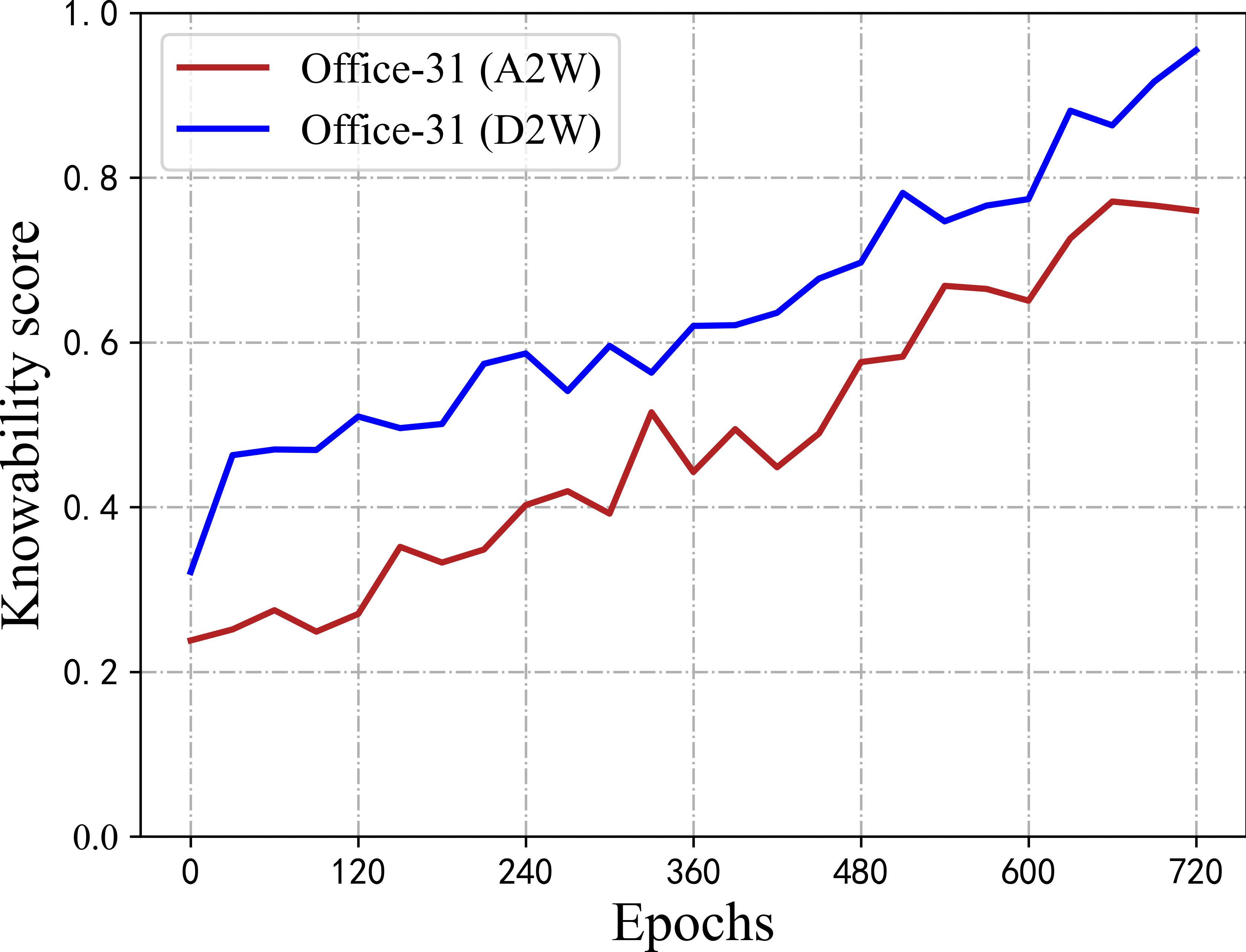}

\caption{ {The knowability score computed as the average of the knowability of all known target samples in one epoch.}}
\label{jjrec}
\end{figure}

{\bf {Justification of the Knowability-Guided Detection.}} To justify the  knowability-guided detection, we visualise the distributions of the knowability of samples on Office-$31$ (A$2$D). As plotted in Fig. \ref{jj}, the distributions of the knowability of the known and the unknown samples have little overlap, which indicates that the unknown samples can be reliably distinguished from known ones by the knowability-guided detection.  {We also conduct experiments to monitor the changes in knowability scores throughout the training process for sub-tasks A$2$W and D$2$W on Office-31 and show the results in Fig. \ref{jjrec}. We can observe that the mean knowability score of known samples consistently increases throughout the training process. It indicates that the inherent distribution of the target samples is progressively becoming more similar to that of the source samples belonging to the same class. Moreover, the increased similarity also indicates that the inter-sample affinity between the source classes and the known target classes becomes higher.} 
 {\textbf{Effect of the Number of Neighbors.} We conduct experiments to explore the influence of different values of $k$ on the $k$-nearest neighbor calculation. As shown in Table \ref{tab:knn}, every dataset has an optimal value of $k$ related to the size of source domain. When $k$ is larger than the optimal value, the performance tends to decrease. Although increasing the value of $k$ moderately can enhance the reliability of the first singular vector, setting $k$ to a large value leads to a significant increase of the noise in the neighborhood, which is influenced by the size of each category in the two domains.  For example, since the subset `Amazon' is three times larger than the subset `Webcam', we can observe that the optimal values in sub-tasks A$2$D and D$2$A are larger than those in D$2$W and W$2$D. Moreover, increasing $k$ will significantly raise the computational cost. But it does not mean that we can always increase $k$ to pursue a performance gain when we have enough computational resources. Therefore, to achieve the optimal performance on average over all sub-tasks and save the computational resources, we select an appropriate value $k = 10$.}

\begin{figure*}[t]
  \centering
   \includegraphics[width=0.9\linewidth]{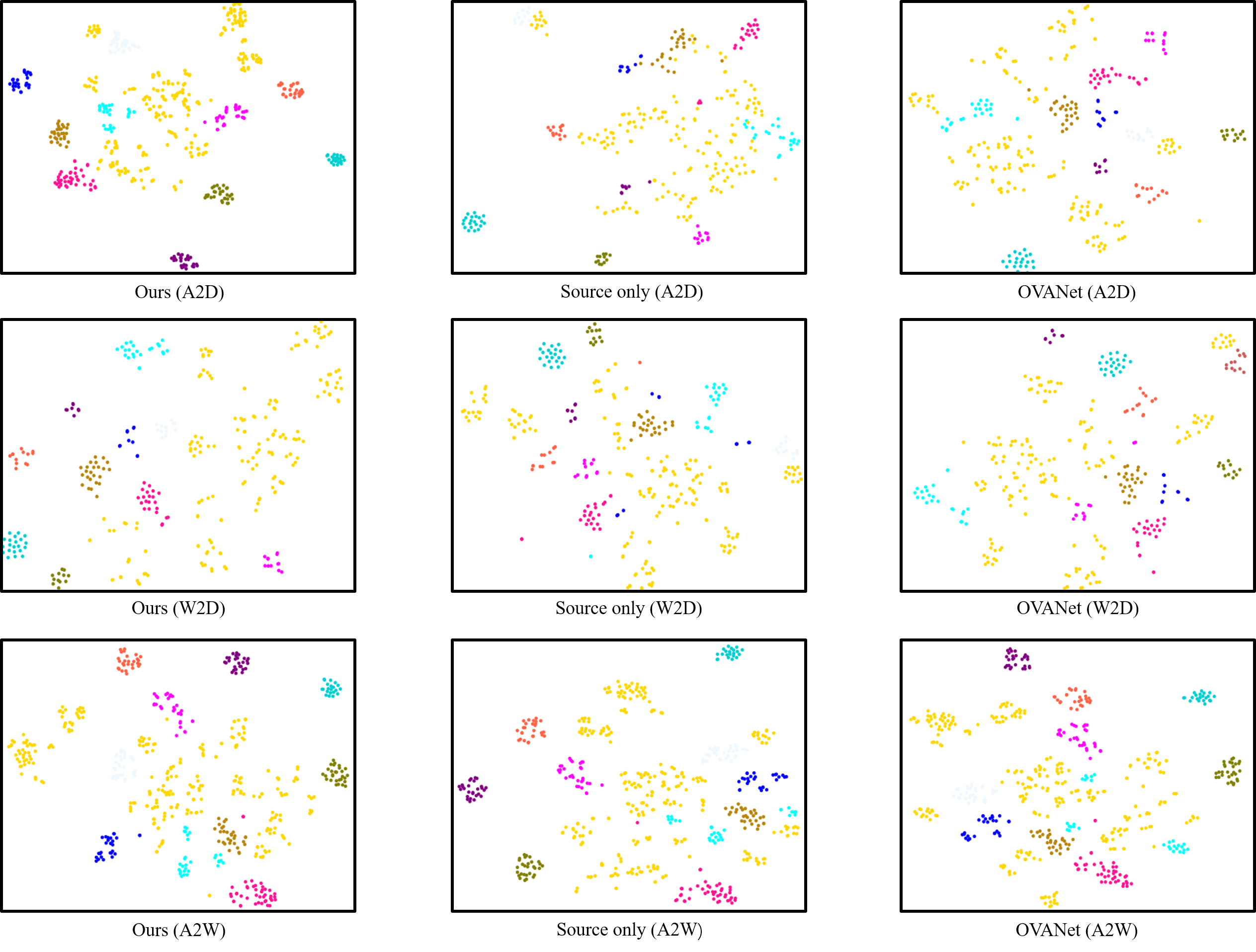}
   \vspace{-1mm}
   \caption{t-SNE visualisations of the classification results produced with different configuration. Different colours represent different classes. Yellow points represent the unknown samples and the points in other colours represent the known samples of different classes. The black dash lines represent the boundaries between the unknown and the known samples while the dash lines in other colours represent the boundaries of the corresponding known classes.}
   \label{tsne}
\end{figure*}

\begin{figure}[t]
\centering
\includegraphics[width=0.9\linewidth]{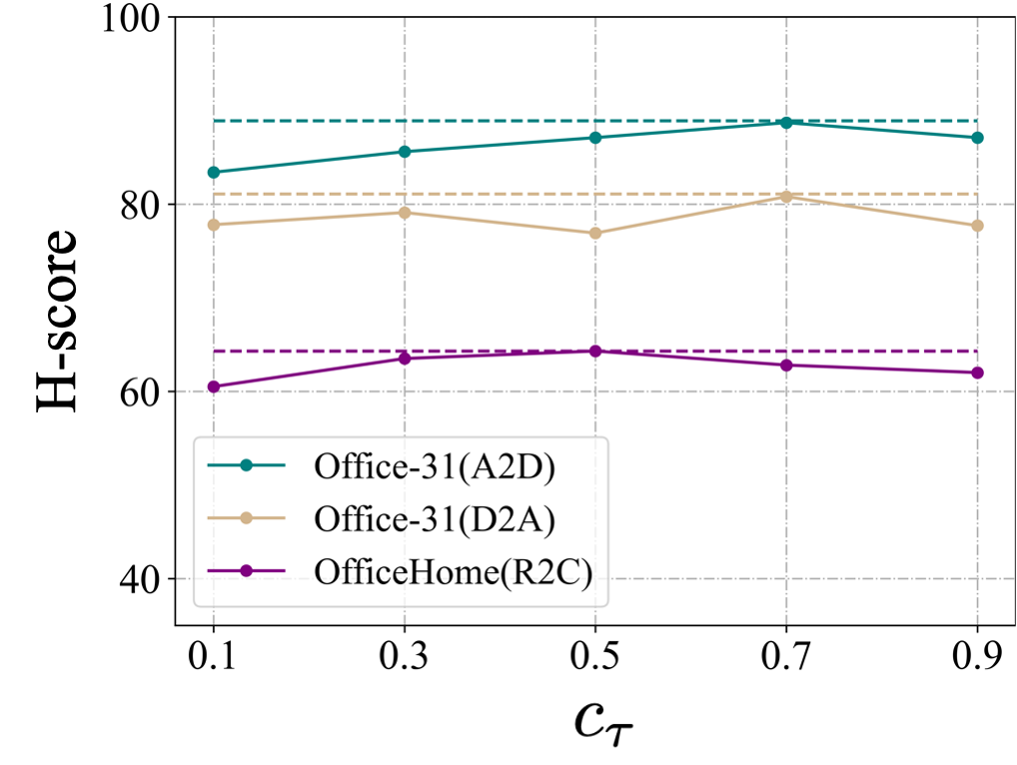}

\caption{Comparison of H-score performance subject to different thresholds where the solid lines represent the results of human-picked thresholds and the dash lines represent the results of the proposed auto-thresholding scheme.}
\label{figctau}
\end{figure}

{\bf Qualitative Comparison by t-SNE Visualisations.} Then, we use t-SNE {\cite{van2008visualizing}} to visualise the features extracted by the feature extractor  $\mathcal{F}(\cdot \mid \phi)$ for the model trained only with the source samples, OVANet, and the proposed method on  Office-31 (A2D, W2D and A2W). As shown in Fig.~\ref{tsne}, before the adaptation to the target domain (middle column), there exists significant misalignment. After the adaptation with the training via OVANet (right column) and our method (left column), the features become more discriminative. We observe better domain alignment as well as target category separation produced by our method. Note that although OVANet does succeed in aligning the source and the target domains and can detect the unknown class, it does not necessarily produce discriminative features for each known class. Moreover, compared to the model trained only with the source samples, the visualisation of our method shows that the inter-sample affinity in each known class increases while that between different classes decreases. 

{\bf Effect of the Auto-Thresholding Scheme.} 
To show the effect of the proposed threshold $c_{\tau}$, we compare it with the human-picked thresholds on Office-$31$ (A$2$D, D$2$A) and OfficeHome (R$2$C). From Fig.~\ref{figctau}, we can observe that it is difficult to choose a consistently optimal threshold for all datasets and sub-tasks as the model is sensitive to the thresholds.

\begin{figure*}[t]
\centering
\subfigure{
\begin{minipage}[t]{0.32\linewidth}
\centering
\includegraphics[width=2.2in]{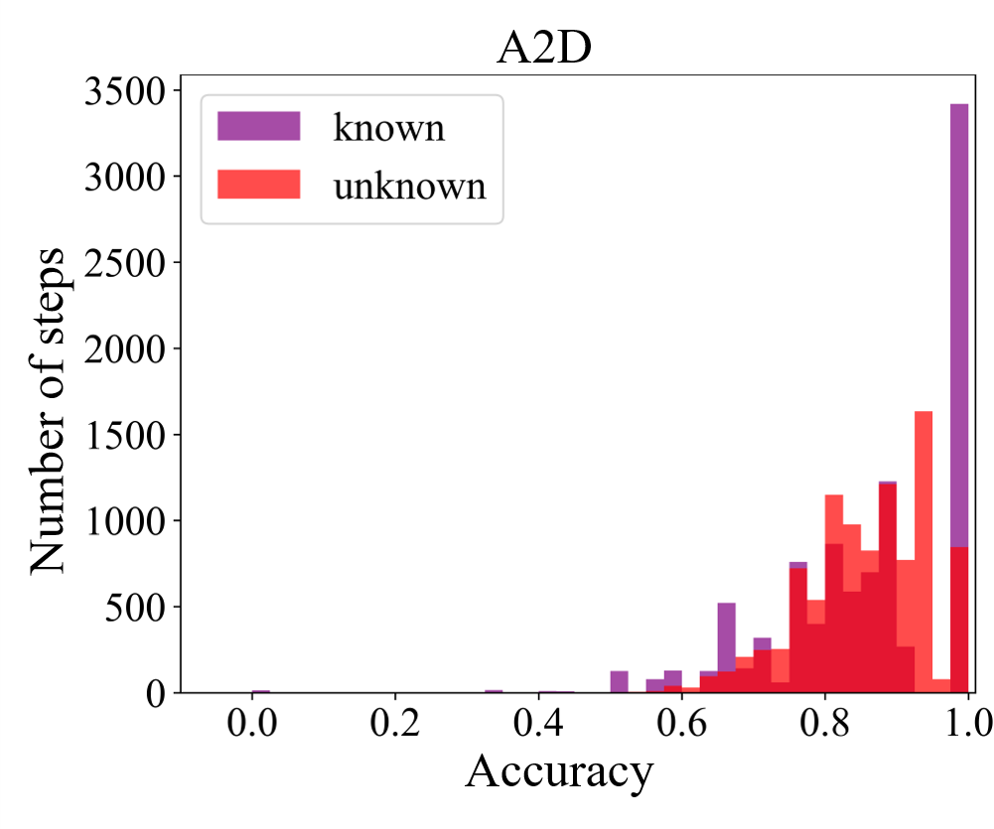}
\end{minipage}%
}%
\subfigure{
\begin{minipage}[t]{0.32\linewidth}
\centering
\includegraphics[width=2.2in]{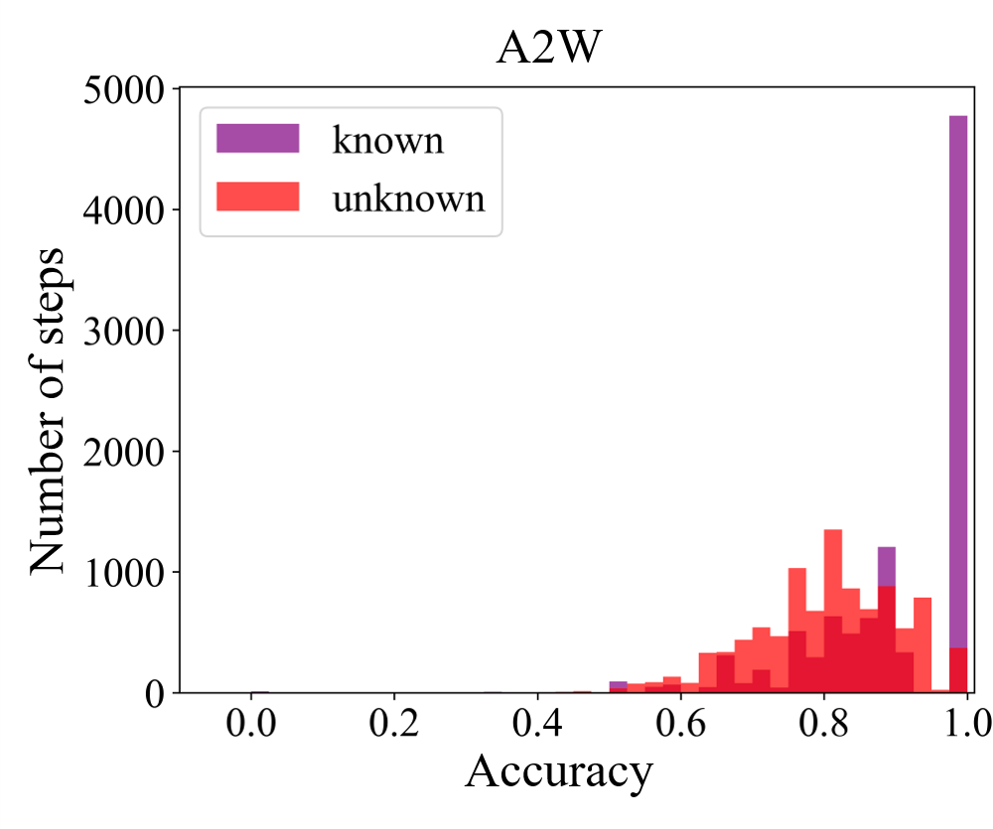}
\end{minipage}%
}%
\subfigure{
\begin{minipage}[t]{0.32\linewidth}
\centering
\includegraphics[width=2.2in]{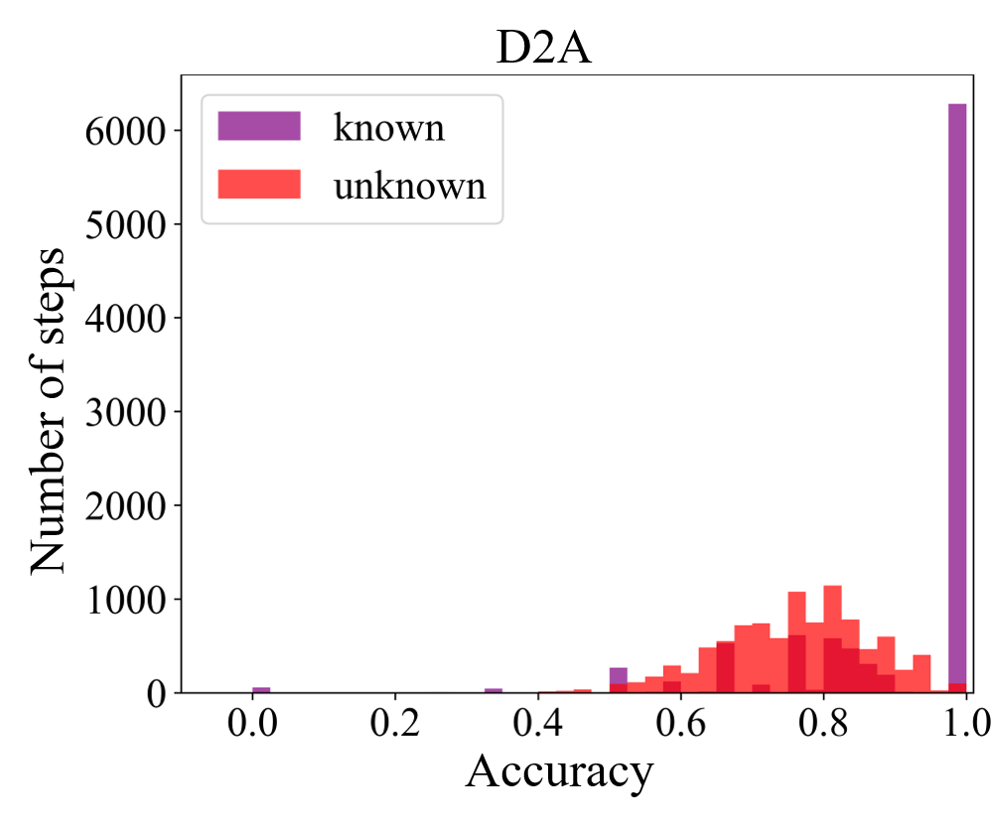}
\end{minipage}%
}%

\subfigure{
\begin{minipage}[t]{0.32\linewidth}
\centering
\includegraphics[width=2.2in]{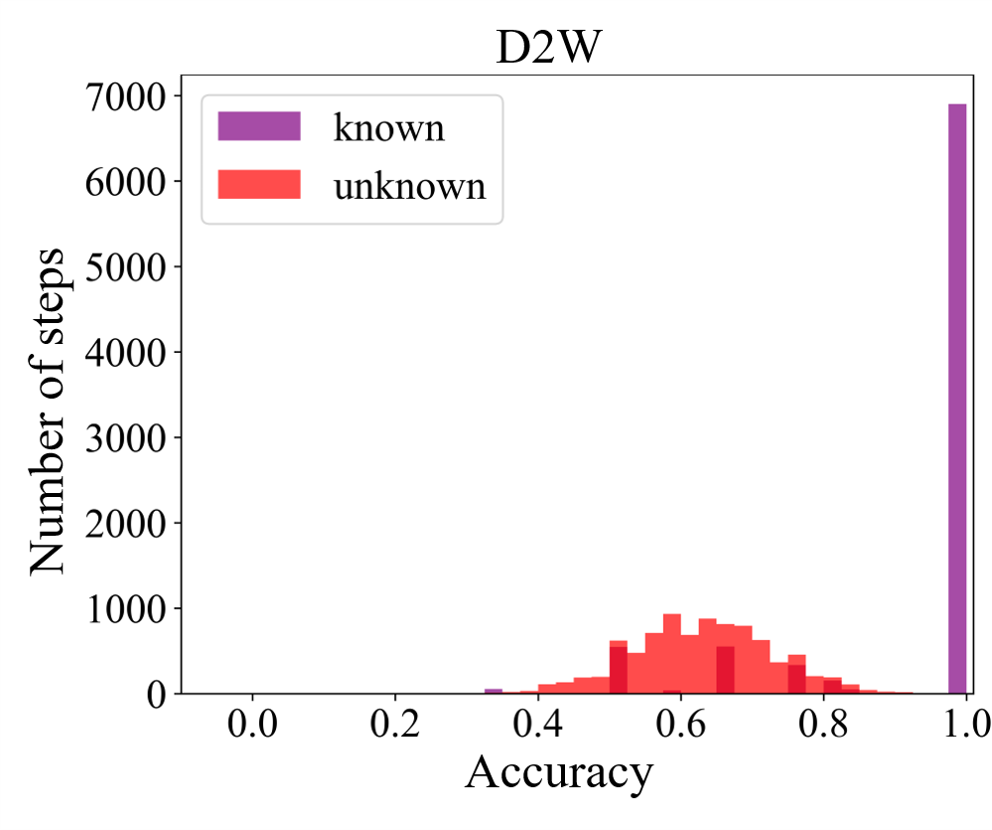}
\end{minipage}%
}%
\subfigure{
\begin{minipage}[t]{0.32\linewidth}
\centering
\includegraphics[width=2.2in]{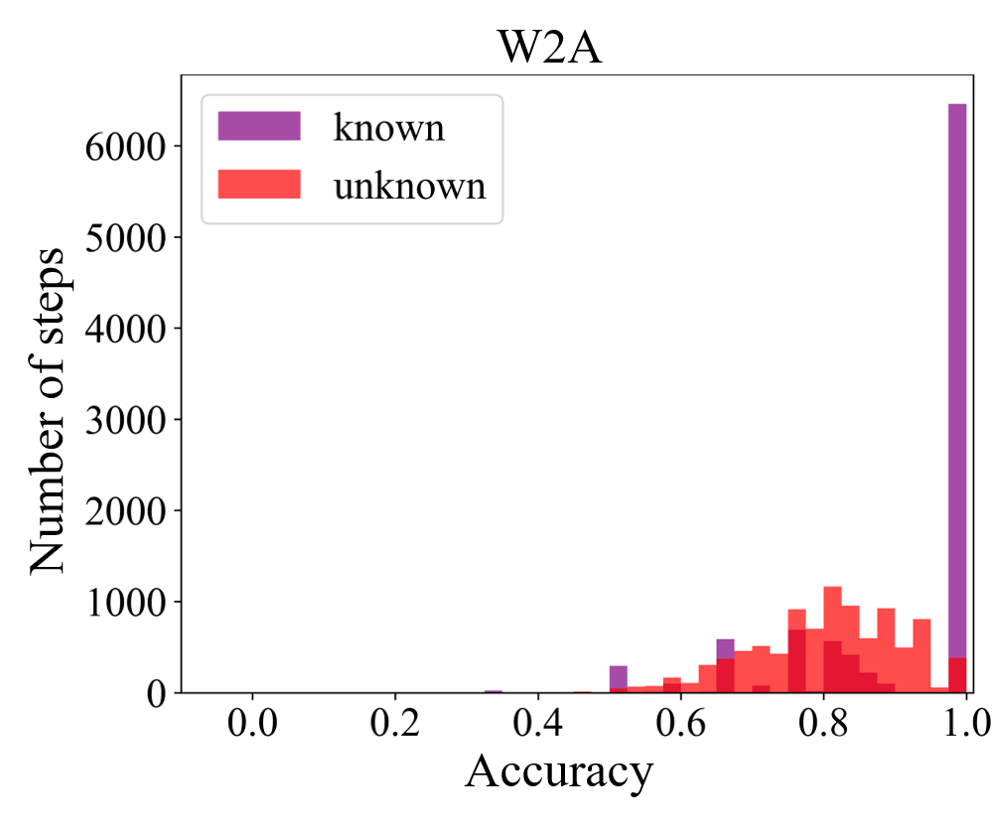}
\end{minipage}%
}%
\subfigure{
\begin{minipage}[t]{0.32\linewidth}
\centering
\includegraphics[width=2.2in]{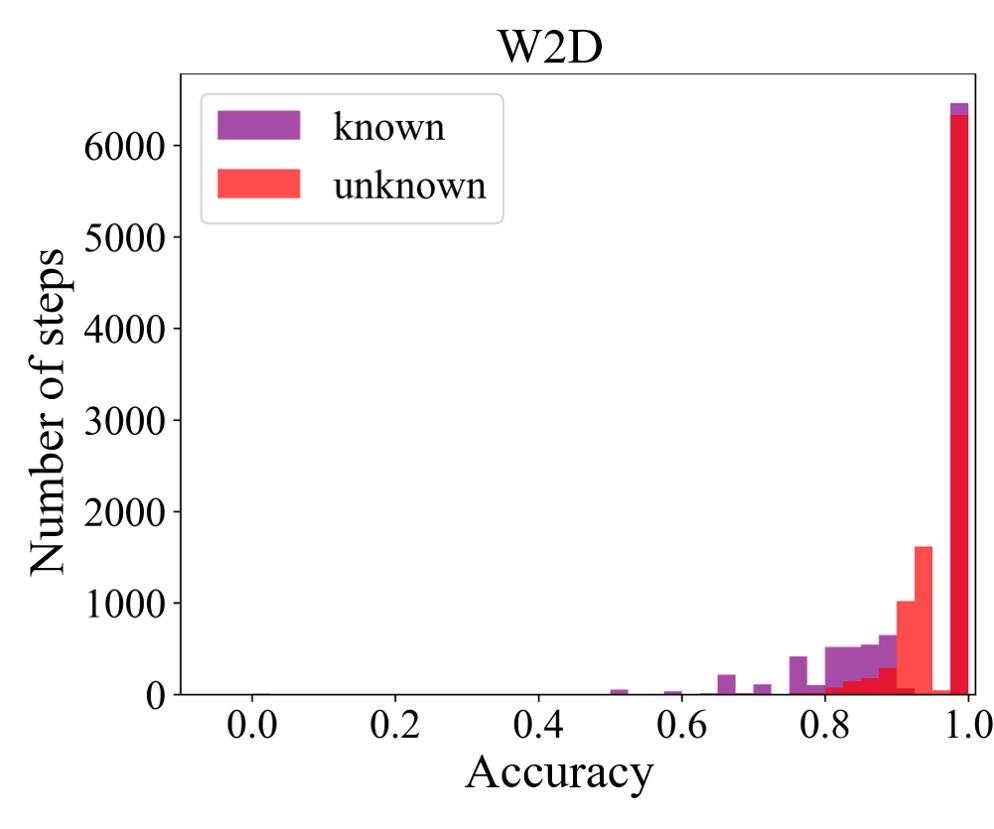}
\end{minipage}%
}%
\centering
\caption{ {\bf Accuracy of the label refinement.}  {Each plot is a histogram illustrating the number of steps at which a particular accuracy of detecting known or unknown samples is achieved.} Plots from left to right in the top row correspond to the sub-tasks A2D, A2W and D2A on Office-$31$, respectively. Plots from left to right in the bottom row correspond to the sub-tasks D2W, W2A and W2D on Office-$31$, respectively. }
\label{acc}
\end{figure*}

\begin{figure*}[t]
\subfigure[$\lambda=\lambda_{k}=\lambda_{unk}=\lambda_{unc}$]{
\begin{minipage}[t]{0.24\linewidth}
\centering
\includegraphics[width=1.6in]{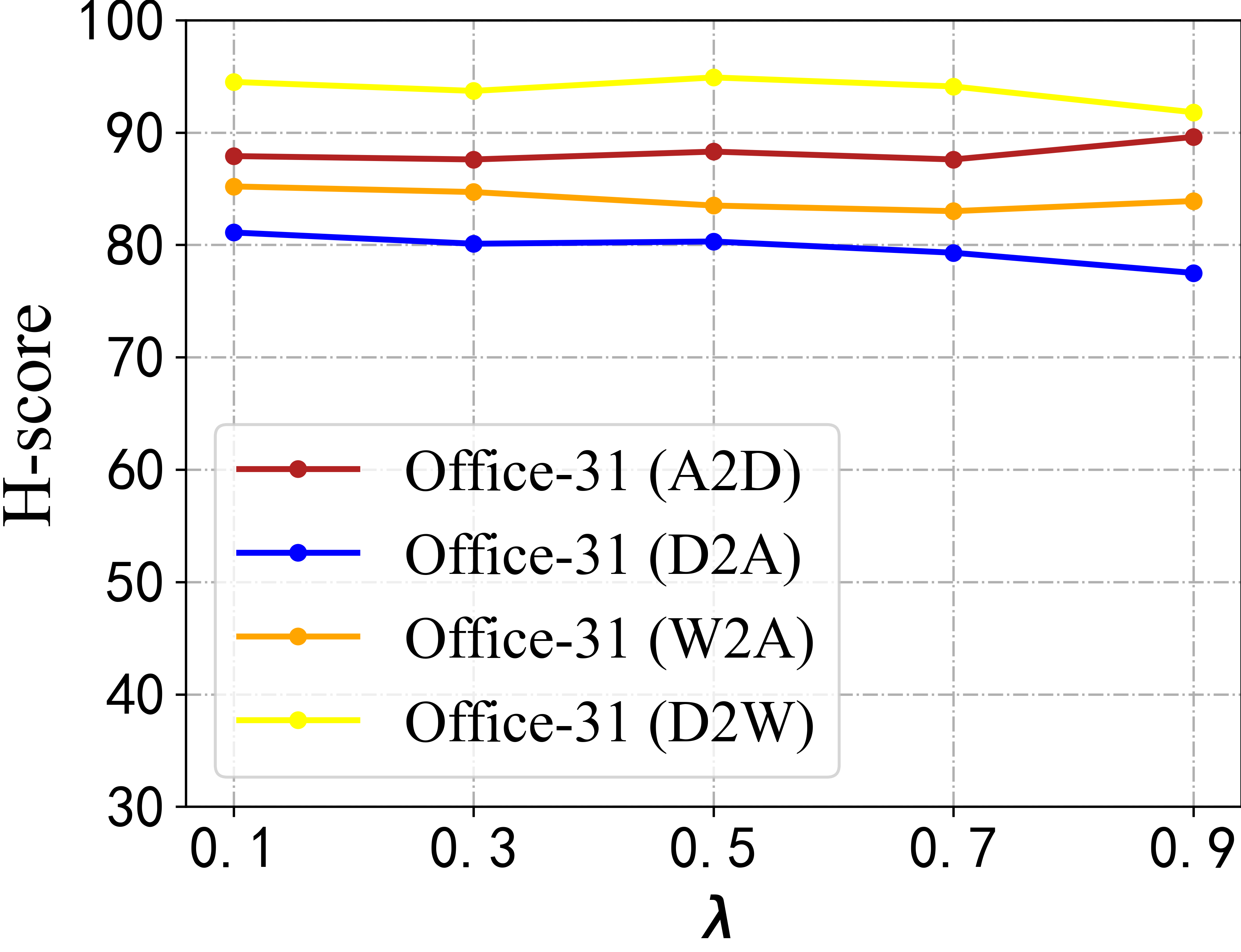}
\end{minipage}%
}%
\hspace{-1mm}
\subfigure[$\lambda_{k}|\lambda_{unk}=\lambda_{unc}=0.1$]{
\begin{minipage}[t]{0.24\linewidth}
\centering
\includegraphics[width=1.6in]{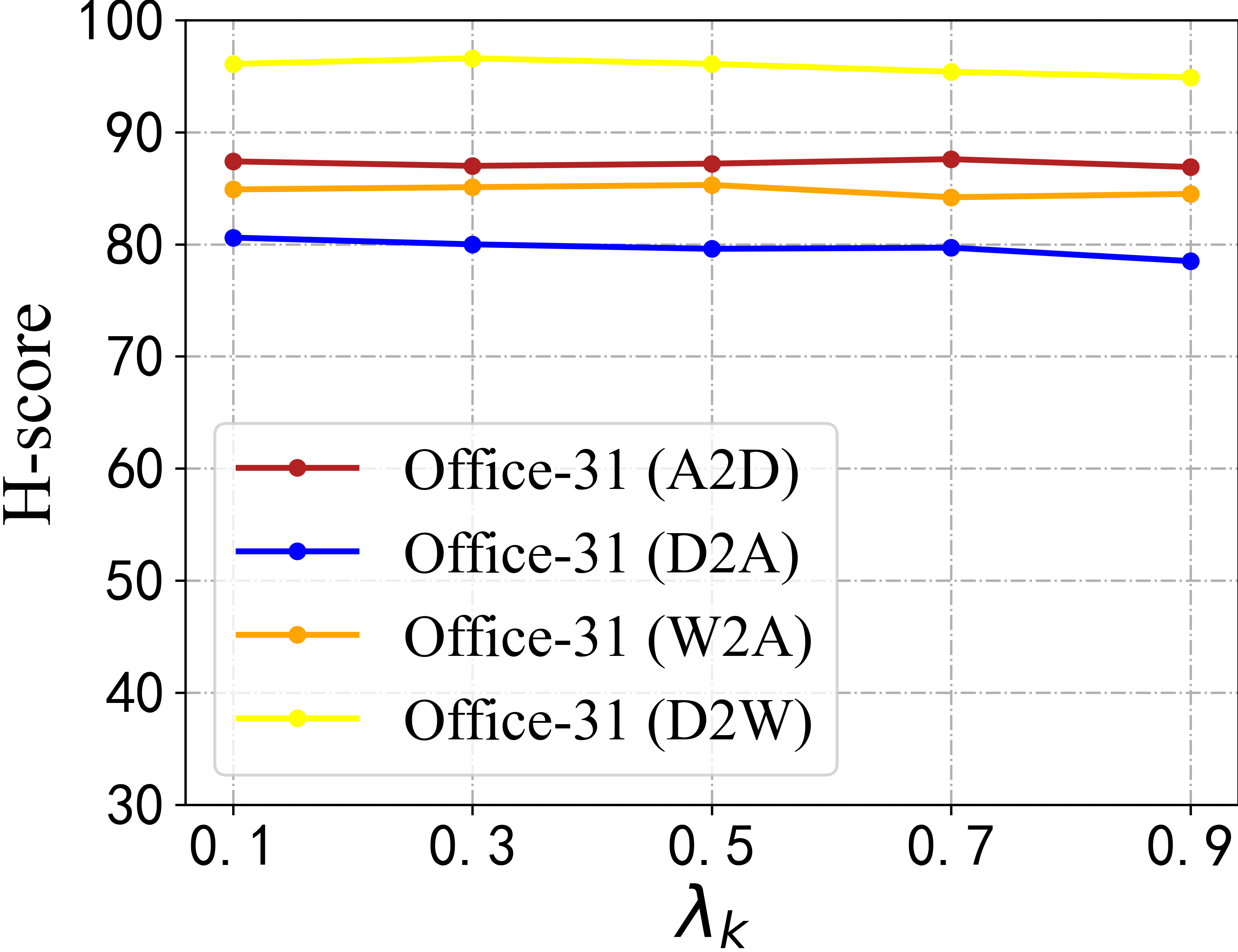}
\end{minipage}%
}%
\hspace{-1mm}
\subfigure[$\lambda_{unk}|\lambda_{k}=\lambda_{unc}=0.1$]{
\begin{minipage}[t]{0.24\linewidth}
\centering
\includegraphics[width=1.6in]{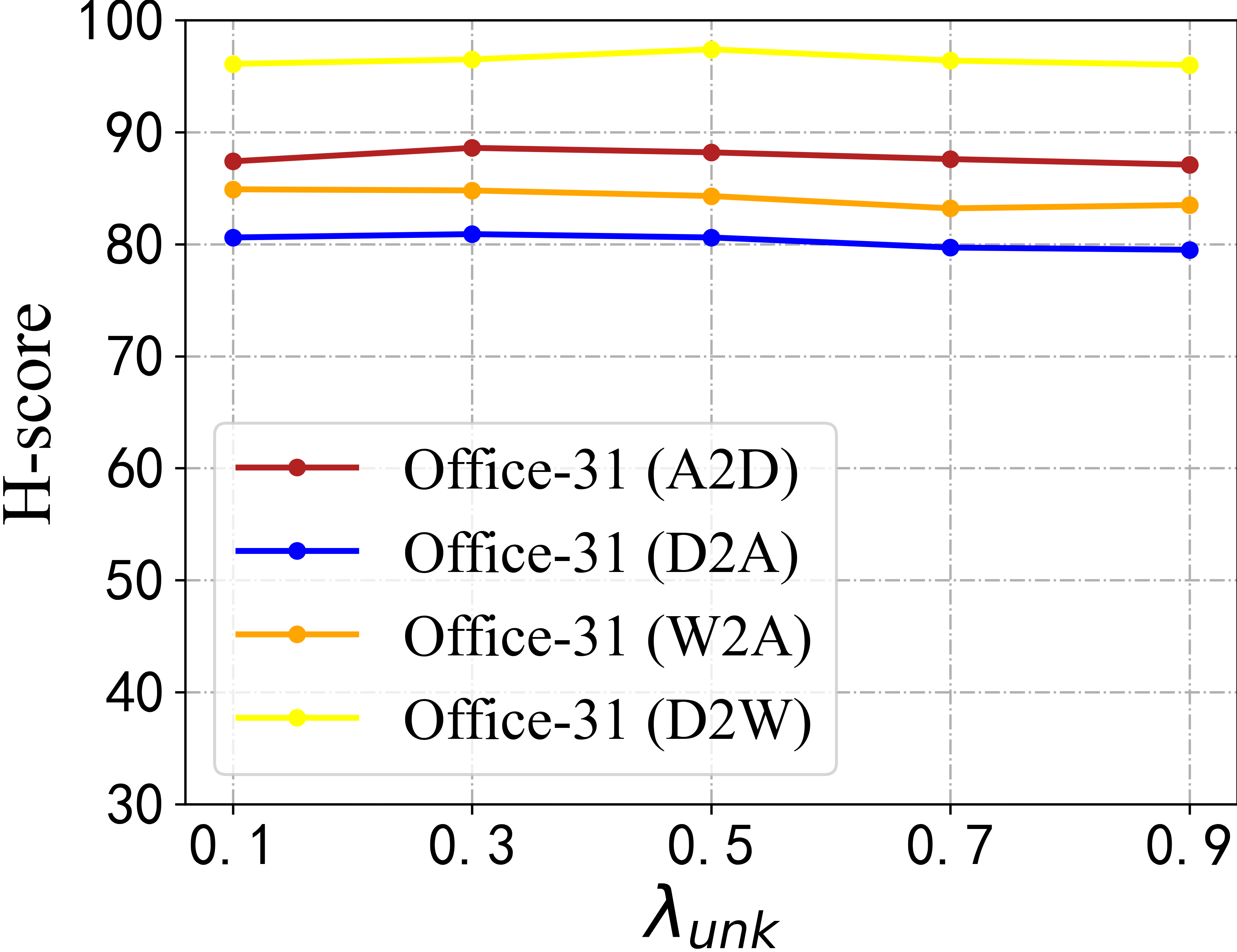}
\end{minipage}%
}%
\hspace{-1mm}
\subfigure[$\lambda_{unc}|\lambda_{k}=\lambda_{unk}=0.1$]{
\begin{minipage}[t]{0.24\linewidth}
\centering
\includegraphics[width=1.6in]{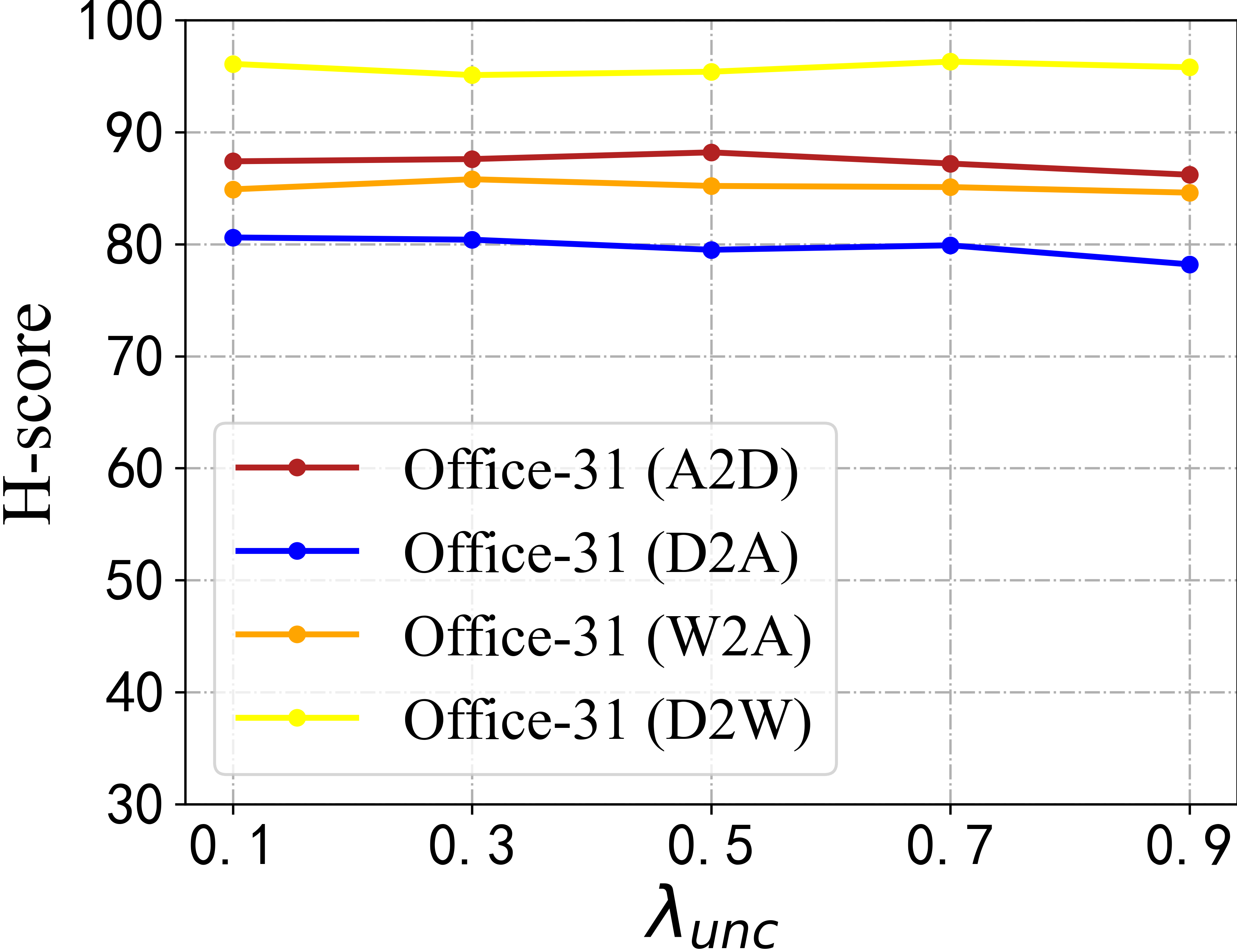}
\end{minipage}%
}%

\centering
\caption{ {Sensitivity to $\lambda$, $\lambda_{k}$, $\lambda_{unk}$ and $\lambda_{unc}$ in terms of H-score. \textbf{ (a)} We show the results with different values of $\lambda$ where we set $\lambda$, $\lambda_{k}$, $\lambda_{unk}$ and $\lambda_{unc}$ all the same. \textbf{ (b), (c) and (d)} We set $\lambda_{k}$, $\lambda_{unk}$ and $\lambda_{unc}$ separately and the results show that our model has a stable performance on different testing sub-tasks.}}
\label{lambdasep}
\end{figure*}

\begin{figure*}[t]
  \centering
   \includegraphics[width=0.9\linewidth]{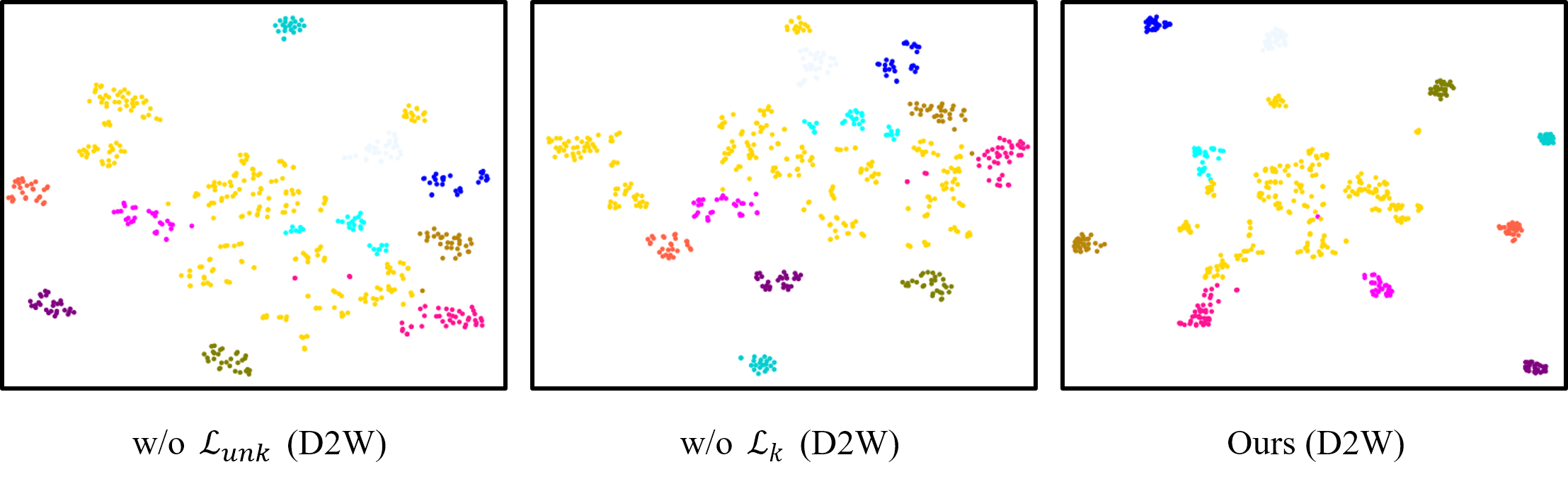}
   \caption{t-SNE visualisations on Office-31 (D2W). Different colors represent different classes. Yellow points represent the unknown samples and the points in other colours represent the known samples of different classes.}
   \label{tsne_loss}
\end{figure*}

{\bf Accuracy of KLS.} We conduct experiments on all sub-tasks of Office-31 where we record the accuracy of KLS for detecting the known/unknown samples at different training steps.  {As plotted in Fig. \ref{acc}, each bar in the histogram represents the number of steps at which a particular accuracy  of detecting known or unknown samples is achieved. For example, the top left plot with regard to the sub-task A$2$D on Office-31 shows that the number of steps at which the accuracy of known samples achieves $1$ is approximately $3,500$. Moreover, there are approximately $1,000$ steps, where the accuracy of unknown samples achieves $1$.}
In Fig.~\ref{acc}, the labeling scheme is consistently estimated with high accuracy which far surpasses $0.7$ on average for both known and unknown samples. Thus, through the proposed knowability-based labeling scheme, our approach reliably finds the unknown and the known samples in the target domain.

{\bf Sensitivity of the Hyper-parameter $\lambda$.} There is only one hyper-parameter $\lambda$ in our model. 
To show the sensitivity of $\lambda$ in the total loss, we conducted experiments on Office-31 with the UniDA setting. Please note the scale of  $\mathcal{L}_{unk} + \mathcal{L}_{k} + \mathcal{L}_{unc}$  is usually much bigger than $\mathcal{L}_s$ because the training on source samples is supervised.
Fig.~\ref{lambdasep} (a) shows that our method has a highly stable performance over different values of $\lambda$.  {To further demonstrate the effect of each loss functions, we replace $\lambda$ with $\lambda_{unk}$, $\lambda_{k}$, and $\lambda_{unc}$ as follows:}
\begin{equation}
     {\mathcal{L}_{all} = \mathcal{L}_s +  \lambda_{unk}\mathcal{L}_{unk} + \lambda_{k}\mathcal{L}_{k} + \lambda_{unc}\mathcal{L}_{unc}.}
\end{equation}
 {We conduct experiments where the  hyper-parameters $\lambda_{unk}$, $\lambda_{k}$, and $\lambda_{unc}$ are set separately and show the results in Fig. \ref{lambdasep} \textbf{(b)}, \textbf{(c)} and \textbf{(d)}. It can be seen that our method is not sensitive to the change of the hyper-parameters  $\lambda_{unk}$, $\lambda_{k}$, and $\lambda_{unc}$. Thus, we just set them all the same.}

{\bf  Effect of the Proposed Losses.}We provide an ablation study to investigate the effect of each loss in our UniDA framework and show the results in Table \ref{tab:abl}. We can see that all losses contribute to the improvement of the results. In particular, among the three target-domain losses, $\mathcal{L}_{unk}$ has the largest impact on the final performance, which demonstrates that it is very important to reduce the inter-sample affinity between the unknown samples and the known ones.

\begin{table}[t]
    \centering
\caption{{Results of different ablated versions of our method on Office-31.}}
   \setlength{\tabcolsep}{1.2mm}
   {
   \begin{tabular}{c|cccccc|c}
   \hline
    \multirow{2}*{ Method } & \multicolumn{6}{c|}{ Office-31 $(10 / 10 / 11)$} & \multicolumn{1}{l}{} \\
    & A2D & A2W & D2A & D2W & W2D & W2A & Avg \\
    \hline 
     w/o $\mathcal{L}_{s}$ & $29. 2$ & $33.4$ & $31.3$ & $52.5$ & $44.2$ & $27.9$ & $36.4$\\
     w/o $\mathcal{L}_{unk}$ & $81.0$ & $77.5$ & $78.2$ & $9 5 . 0$ & $91.0$ & $72.9$ & $82.6$ \\

     w/o $\mathcal{L}_{unc}$ & $8 6 . 9$ & $76.6$ & $8 4.4$ & $91.4$ & $93.3$ & $85.6$ & $8 6 . 3$\\
     w/o $\mathcal{L}_{k}$ & $86.2$ & $80.6$ & $79.5$ & $93.9$ & $97.5$ & $81.8$ & $86.5$ \\
    \hline
    Ours  & $8 9 . 5$ & $8 4 . 9$ & $8 9 . 7$ & $93.7$ & $85.8$ & $8 8 . 5$ & $8 8 . 7$\\
    \hline
    \end{tabular}}
    \label{tab:abl}
\end{table}

To further show the effect of the proposed losses, we use t-SNE algorithm to visualise the features of target samples on Office-31 (D2W). As plotted in Fig.~\ref{tsne_loss}, without $\mathcal{L}_{unk}$ (left), the boundary between the unknown and the known samples is unclear. Without $\mathcal{L}_{k}$ (middle), samples belonging to a known class are not compact. However, the inter-sample affinity between the unknown and the known samples produced by the full version of our method (right), is much lower than that produced without $\mathcal{L}_{unk}$. And the inter-sample affinity in a known class produced by the full version of our method is much higher than that produced without $\mathcal{L}_{k}$. Such results demonstrate the main idea of the proposed method.

\begin{figure*}[t]
  \centering
   \includegraphics[width=0.9\linewidth]{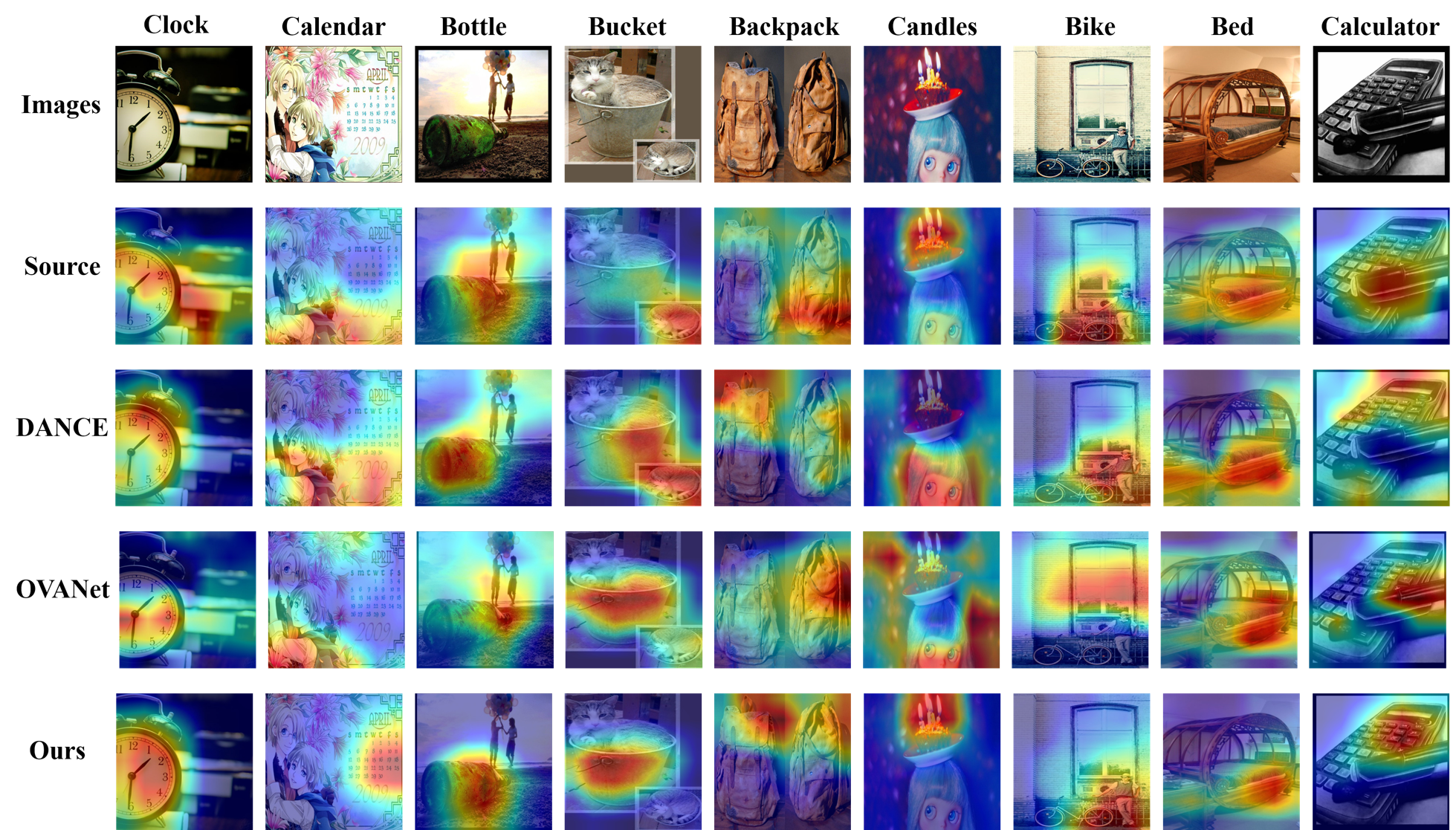}
   \caption{Grad-CAM \cite{selvaraju2017grad} visualisations of different methods on the sub-task R2A of OfficeHome. Generally, our method shows good concentration on known target samples and focuses on a variety of relevant regions.}
   \label{heatmapde}
\end{figure*}

{\bf Sensitivity of scales for $c_{\tau}$.}
Instead of using an automatic scheme, we set the parameter $c_{\tau}$ to $0.8$ empirically. This is because changing $c_{\tau}$ has little influence on the performance. To verify this point, we test our method with different $c_{\tau}$ and show the results in Table~\ref{tab:ctau}.

\begin{table}[t]
\setlength\tabcolsep{2mm}
   \centering
       \caption{Results on Office-31 with different scales of $c_{\tau}$ under the UniDA setting (H-score).}
   \setlength{\tabcolsep}{2.1mm}
\resizebox{0.49\textwidth}{14mm}
   {
   \begin{tabular}{l|cccccc|c}
   \hline { } & \multicolumn{6}{c|}{ Office-31 $(10 / 10 / 11)$} & \multicolumn{1}{l}{} \\
    & A2D & A2W & D2A & D2W & W2D & W2A & Avg \\
    \hline $0.1c_{\tau}$ & $84.6$ & $81.7$ & $83.2$ & $94.0$ & $97.5$ & $\mathbf{86.1}$ & $87.9$ \\
    $0.3c_{\tau}$ & $86.3$ & $81.2$ & $\mathbf{82.6}$ & $\mathbf{95.0}$ & $97.4$ & $85.0$ & $87.9$\\
    $0.5c_{\tau}$ & $88.1$ & $\mathbf{83.3}$ & $81.7$ & $94.4$ & $97.0$ & $85.4$ & $88.3$\\
    $0.7c_{\tau}$  & $88.4$ & $82.9$ & $81.0$ & $93.8$ & $97.9$ & $84.8$ & $88.1$\\
    $0.8c_{\tau}$  & $\mathbf{88.9}$ & $83.0$ & $81.1$ & $94.5$ & $98.3$ & $85.2$ & $\mathbf{88.4}$ \\
    $0.9c_{\tau}$  & $88.8$ & $83.1$ & $80.3$ & $94.5$ & $\mathbf{99.2}$ & $84.1$ & $8 8.3$ \\

   \hline
    \end{tabular}}
           \setlength{\abovecaptionskip}{1pt}%
    \setlength{\belowcaptionskip}{7pt}%

       \label{tab:ctau}

  \end{table}

{\bf Visual Explanations with Grad-CAM.}
 In this section, we utilise the visualisation technique Grad-CAM in \cite{selvaraju2017grad} to visualise the predictions and compare the Grad-CAM visualisations \cite{selvaraju2017grad} for different methods in Fig. \ref{heatmapde}. 
To verify the validity of our method, we also visualise the
previous methods including the source only model (second row) and DANCE \cite{saito2020dance} (third row) as well as OVANet \cite{Saito_2021_ICCV} (fourth row) on their predictions. Obviously, we can observe that the semantic capabilities of our method (fifth row) are significantly stronger than OVANet \cite{Saito_2021_ICCV} and DANCE \cite{saito2020dance}. We can also notice that our method concentrates on more relevant regions and the features of principal regions are accentuated, which verifies that our method indeed achieves an improvement for the critical parts in classification. The main reason is that our model learns discriminative information from each part and captures diverse relevant regions, while DANCE \cite{saito2020dance} and OVANet \cite{Saito_2021_ICCV} are usually distracted by and even focus on some irrelevant area.

{\bf  Performance of Using VGGNet as Backbone.} Table~\ref{tab:vgg} shows the quantitative comparison with the ODA setting on Office-31 using ~VGGNet \cite{simonyan2014very} ~instead of  ResNet50 as the backbone for feature extraction. According to the results, we demonstrate that our method is also effective with another backbone without changing any hyper-parameters.

\begin{table}[t] 
    \centering
      \caption{Results on Office-31 using the VGGNet~\cite{simonyan2014very} backbone with the ODA setting.}
   \setlength{\tabcolsep}{1.1mm}
{
    \begin{tabular}{l|cccccc|c}
   \hline
    \multirow{2}*{ Method } & \multicolumn{6}{c|}{ Office-31 $(10 / 10 / 11)$} & \multicolumn{1}{l}{} \\
    & A2D & A2W & D2A & D2W & W2D & W2A & Avg \\
    \hline OSBP \cite{saito2018open} & $81.0$ & $77.5$ & $78.2$ & $\mathbf{9 5 . 0}$ & $91.0$ & $72.9$ & $82.6$ \\
    ROS \cite{bucci2020effectiveness} & $79.0$ & $81.0$ & $78.1$ & $94.4$ & $\mathbf{9 9 . 7}$ & $74.1$ & $84.4$ \\
    OVANet \cite{Saito_2021_ICCV} & $\mathbf{89.5}$ & $\mathbf{84.9}$ & $89.7$ & $93.7$ & $85.8$ & $88.5$ & $88.7$\\
    \hline
 Ours & $89.5$ & ${84 . 6}$ & $\mathbf{92.0}$ & $94.5$ & $91.5$ & $\mathbf{91.8}$ & $\mathbf{90.6}$ \\
   \hline
    \end{tabular}}
    \label{tab:vgg}
\end{table}

\section{Conclusions}
In this paper, we propose a new framework to explore the inter-sample affinity for UniDA. Its core idea is to reduce the inter-sample affinity between the unknown and the known samples while increasing that within the known samples by estimating the knowability of each sample. Extensive experiments demonstrate that our method achieves the SOTA performance in various sub-tasks on four public datasets.

A limitation of our method is that it does not sufficiently utilise the inter-sample relationship within the set of unknown samples. Thus in the future work, we plan to extend our method to leverage this relationship for further 
boosting the performance with the UniDA setting.   {Moreover, since the proposed method assumes that the local affinity distributions of source and target samples within the same class are similar, we will explore the scenario where the distribution of samples in a known category is heterogeneous and differs between source and target domains in the future work.}\\
\vspace{5pt}
\\
{\bf Data Availability Statement}\vspace{5pt}\\ 
The datasets generated during and/or analysed during the current study are available in the Office-31 repository \href{https://faculty.cc.gatech.edu/~judy/domainadapt/}{[\underline{Link}]}, the OfficeHome repository \href{https://www.hemanthdv.org/officeHomeDataset.html}{[\underline{Link}]}, the VisDA repository \href{https://github.com/VisionLearningGroup/taskcv-2017-public/tree/master/classification}{[\underline{Link}]}, and the DomainNet repository \href{http://ai.bu.edu/M3SDA/}{[\underline{Link}]}. \\
\vspace{5pt}
\\
{\bf Declarations}\vspace{5pt}\\ 
The authors have no relevant financial or non-financial interests to disclose.

\bibliographystyle{splncs04}
\bibliography{egbib}



\end{document}